\documentclass[11pt,twoside,twocolumn,a4paper]{article}

\usepackage{cvww}
\usepackage{times}
\usepackage{epsfig}
\usepackage{graphicx}
\usepackage{amsmath}
\usepackage{amssymb}

\usepackage[usenames,dvipsnames]{color}
\usepackage{xcolor}
\usepackage{booktabs}
\usepackage{subfig}
\usepackage{placeins}
\usepackage{multirow}
\usepackage{varwidth}
\usepackage[font=small,skip=-10pt]{caption}

\cvwwfinalcopy 

\def\cvwwPaperID{2} 

\ifcvwwfinal\fi
\begin{document}

\title{Hands Deep in Deep Learning for Hand Pose Estimation}

\author{Markus Oberweger  \qquad Paul Wohlhart \qquad Vincent Lepetit\\
Institute for Computer Graphics and Vision\\
Graz University of Technology, Austria\\
{\tt\small \{oberweger, wohlhart, lepetit\}@icg.tugraz.at}
}

\maketitle
\ifcvwwfinal\thispagestyle{fancy}\fi

\begin{abstract}

We  introduce  and  evaluate  several  architectures  for  Convolutional  Neural
Networks to  predict the 3D  joint locations  of a hand  given a depth  map.  We
first  show  that  a  prior  on  the  3D  pose  can  be  easily  introduced  and
significantly improves the accuracy and  reliability of the predictions. We also
show how  to use context efficiently  to deal with ambiguities  between fingers.
These   two   contributions   allow   us   to   significantly   outperform   the
state-of-the-art on  several challenging benchmarks,  both in terms  of accuracy
and computation times. The code can be found at \small\url{https://github.com/moberweger/deep-prior/}.

\end{abstract}


\section{Introduction}

Accurate  hand  pose estimation  is  an  important  requirement for  many  Human
Computer  Interaction or  Augmented Reality  tasks,  and has  attracted lots  of
attention         in         the        Computer         Vision         research
community~\cite{Keskin2011,Keskin2012,Melax2013,Oikonomidis2011,Qian2014,Tang2014,Tang2013,Xu2013}.
Even with 3D  sensors such as structured-light or time-of-flight  sensors, it is
still very  challenging, as the hand  has many degrees of  freedom, and exhibits
self-similarity and self-occlusions in images.

Given  the  current trend  in  Computer  Vision, it  is  natural  to apply  Deep
Learning~\cite{Schmidhuber14}  to solve  this task,  and a  Convolutional Neural
Network~(CNN) with  a standard architecture  performs remarkably well when  applied to
this problem, as a simple experiment  shows.  However, the layout of the network
has a strong influence on  the accuracy of the output~\cite{Coates2011,Sun2013}
and in this paper, we aim at  identifying the architecture that performs best for
this problem.

More specifically, our contribution is two-fold:
\begin{itemize}
\item We show that we can learn a  prior model of the hand pose and integrate it
  seamlessly to the network to improve the accuracy of the predicted pose.  This
  results  in a  network  with an  unusual ``bottleneck'', \ie a  layer with  fewer
  neurons  than the  last layer.
\item Like previous work~\cite{Sun2013,Toshev2014}, we use a refinement stage to
  improve the location estimates for each  joint independently. Since it is
  a regression problem, spatial pooling and subsampling should be used carefully for this stage.
  To solve this problem, we use  multiple input regions centered on the initial
  estimates of the joints, with very small pooling regions for the smaller input
  regions, and  larger pooling  regions for the  larger input  regions.  Smaller
  regions provide accuracy, larger regions provide contextual information.
\end{itemize}

We show  that our original  contributions allow us to  significantly outperform
the          state-of-the-art           on          several          challenging
benchmarks~\cite{Tang2014,Tompson2014},   both   in   terms  of   accuracy   and
computation  times.  Our  method runs at over 5000~fps  on a  single GPU  and over
500~fps  on   a  CPU,  which  is   one  order  of  magnitude   faster  than  the
state-of-the-art.

In the remainder of  the paper, we first give a short review  of related work in
Section~\ref{sec:relatedwork}.     We    introduce    our    contributions    in
Section~\ref{sec:main} and evaluate them in Section~\ref{sec:eval}.

\section{Related Work}
\label{sec:relatedwork}

Hand pose estimation is an old problem in Computer Vision, with early references
from the  nineties, but  it is  currently very active probably because of the
appearance of depth sensors.  A good overview of earlier work is given 
in~\cite{Erol2007}. Here we will discuss only more recent work, which can be
divided into two main approaches.

The  first  approach  is  based on  generative,  model-based  tracking  methods.
\cite{Oikonomidis2011,Qian2014}  use   a  3D  hand  model   and  Particle  Swarm
Optimization   to  handle   the  large   number  of   parameters  to   estimate.
\cite{Melax2013} also  considers dynamics simulation  of the 3D  model.  Several
works  rely on  a tracking-by-synthesis  approach: \cite{LaGorce2011}  considers
shading  and texture, \cite{Ballan2012} salient  points, and \cite{Xu2013} depth
images.  All  these works require  careful initialization in order  to guarantee
convergence and  therefore rely on  tracking based on  the last frames'  pose or
separate  initialization  methods---for  example, \cite{Qian2014}  requires  the
fingertips to be visible.  Such  tracking-based methods have difficulty handling
drastic changes between two  frames, which are common as the  hand tends to move
fast.

The second type  of approach is discriminative, and aims  at directly predicting
the  locations  of   the  joints  from  RGB  or  RGB-D   images.   For  example,
\cite{Keskin2012} and \cite{Kuznetsova2013} rely on multi-layered Random Forests
for the  prediction.  The former uses  invariant depth features, and  the latter
uses clustering in hand configuration  space and pixel-wise labelling.  However,
both do not predict  the actual 3D pose but only  classify given poses based
on     a     dictionary.     Motivated     by     work     for    human     pose
estimation~\cite{Shotton2011}, \cite{Keskin2011} uses Random Forests to
perform a per-pixel classification of depth images and then a local mode-finding
algorithm to estimate the 2D  joint locations.  However, this approach cannot
directly infer the locations of hidden joints, which are much more frequent for
hands than for the human body.

\cite{Tang2013}  proposed  a  semi-supervised  regression  forest,  which  first
classifies the hands  viewpoint, then the individual joints,  to finally predict
the  3D   joint  locations.    However,  it  relies   on  a   costly  pixel-wise
classification,  and  requires  a  huge   training  database  due  to  viewpoint
quantization.  The same authors  proposed a regression forest in~\cite{Tang2014}
to directly regress  the 3D locations of the joints,  using a hierarchical model
of the hand.   However, their hierarchical approach  accumulates errors, causing
larger errors for the finger tips.

Even more  recently, \cite{Tompson2014} uses  a CNN for  feature extraction
and generates small ``heatmaps'' for joint   locations from which they infer the
hand pose using  inverse kinematics.  However, their approach  predicts only the
2D locations of the joints, and uses a depth map for the third coordinate, which
is problematic  for hidden joints.   Furthermore, the accuracy is  restricted to
the heatmap resolution, and creating  heatmaps is computationally costly as the
CNN has to be evaluated at each pixel location.

The hand pose estimation problem is of  course closely related to the human body
pose estimation  problem.  To  tackle this problem,  \cite{Shotton2011} proposed
per-pixel semantic segmentation and regression  forests to estimate the 3D human
body pose from a single depth  image.  \cite{Ionescu2014} recently showed it was
possible to do  the same from RGB  images only, by combined  body part labelling
and   iterative  structured-output   regression  for   3D  joint   localization.
\cite{Toshev2014} recently proposed  a cascade of CNNs  to directly predict
and  iteratively  refine  the  2D  joint  locations  in  RGB  images.   Further,
\cite{Tompson2014a}  used a  CNN for  part detection  and a  simple spatial
model, which however, is not effective for high variations in pose space.

In  our work,  we build  on the  success  of CNNs  and use  them for  their
demonstrated performance. We observe, that the  structure of the network is very
important. Thus we  propose and investigate different architectures  to find the
most appropriate one for the hand pose estimation problem.  We propose a network
structure that  works very  well, outperforming the  baselines on  two difficult
datasets.


\section{Hand Pose Estimation with Deep Learning}

\label{sec:main}

In  this  section we  present  our  original  contributions  to the  hand  pose
estimation problem.  We first briefly introduce the problem and a simple 2D hand
detector, which we use to get a coarse  bounding box of the hand as input to the
CNN-based pose predictors.

Then we  describe our  general approach  which consists of  two stages.  For the
first stage  we consider different  architectures that predict the  locations of
all joints  simultaneously. Optionally,  this stage  can predict  the pose  in a
lower-dimensional space, which is described  next. Finally, we detail the second
stage,  which  refines  the  locations  of the  joints  independently  from  the
predictions made at the first stage.

\subsection{Problem Formulation}

We  want   to  estimate  the   $J$  3D   hand  joint  locations   $\mathbf{J}  =
\{\mathbf{j}_i\}^J_{i=1}$  with $\mathbf{j}_i  =  (x_i,y_i,z_i)$ from  a single  depth
image.  We assume that a training set of depth images labeled with
the 3D joint locations is available.   To simplify the regression task, we first
estimate a  coarse 3D  bounding box  containing the hand  using a  simple method
similar to  \cite{Tang2014}, by assuming the  hand is the closest  object to the
camera: We extract from  the depth map a fixed-size cube  centered on the center
of mass of this object, and resize it to a $128\times128$ patch of
depth  values  normalized to  $[-1,1]$.   Points  for  which  the depth  is  not
available---which may happen with structured  light sensors for example---or are
deeper  than the  back  face of  the  cube, are  assigned a  depth  of 1.   This
normalization  is  important for  the  CNN  in  order  to be  invariant  to
different distances from the hand to the camera.

\begin{figure*}
\begin{center}
\subfloat[]{
   \includegraphics[width=0.32\textwidth]{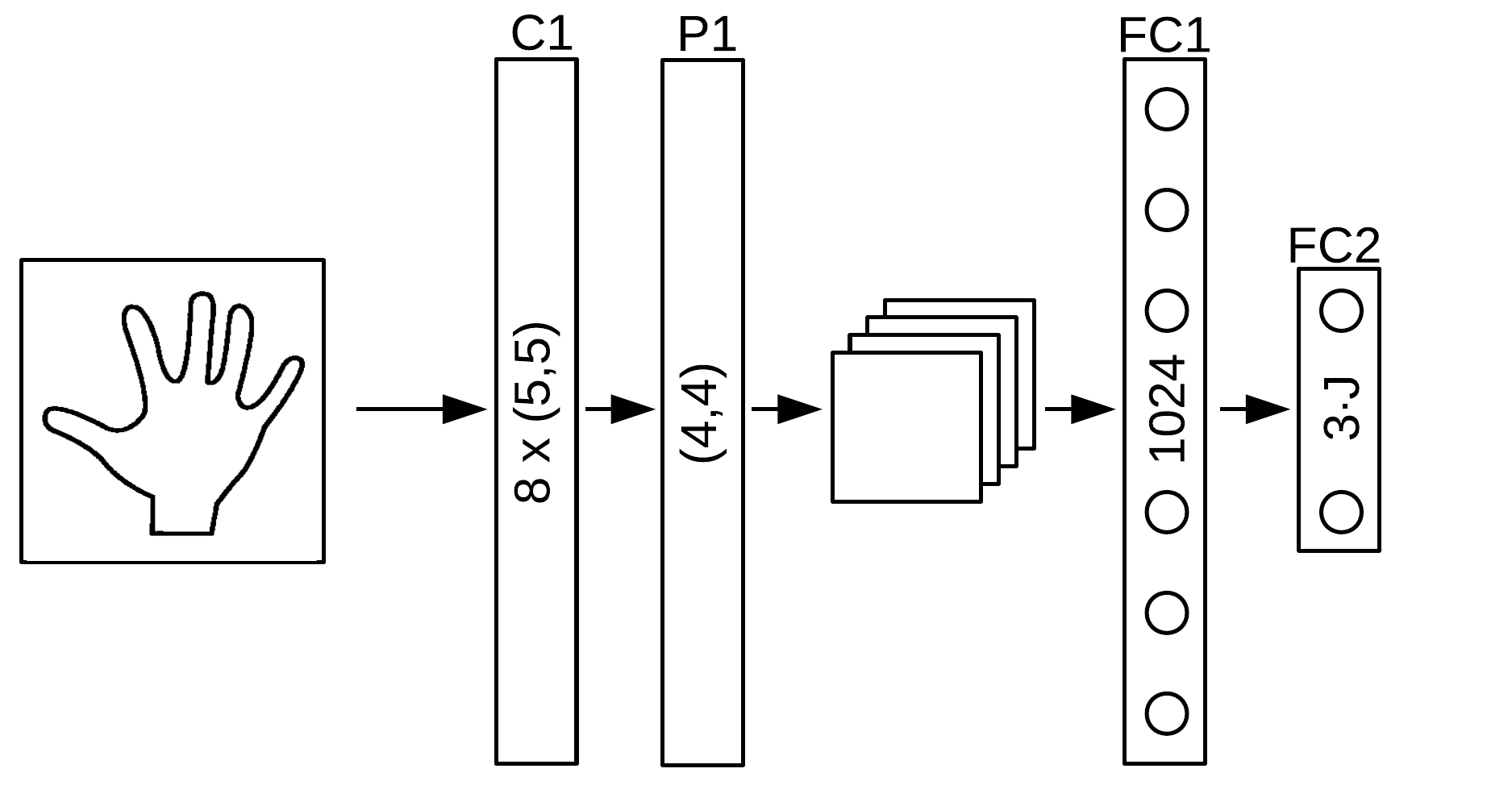}\label{fig:shallownet}
}\hspace{1cm}
\subfloat[]{
   \includegraphics[width=0.47\textwidth]{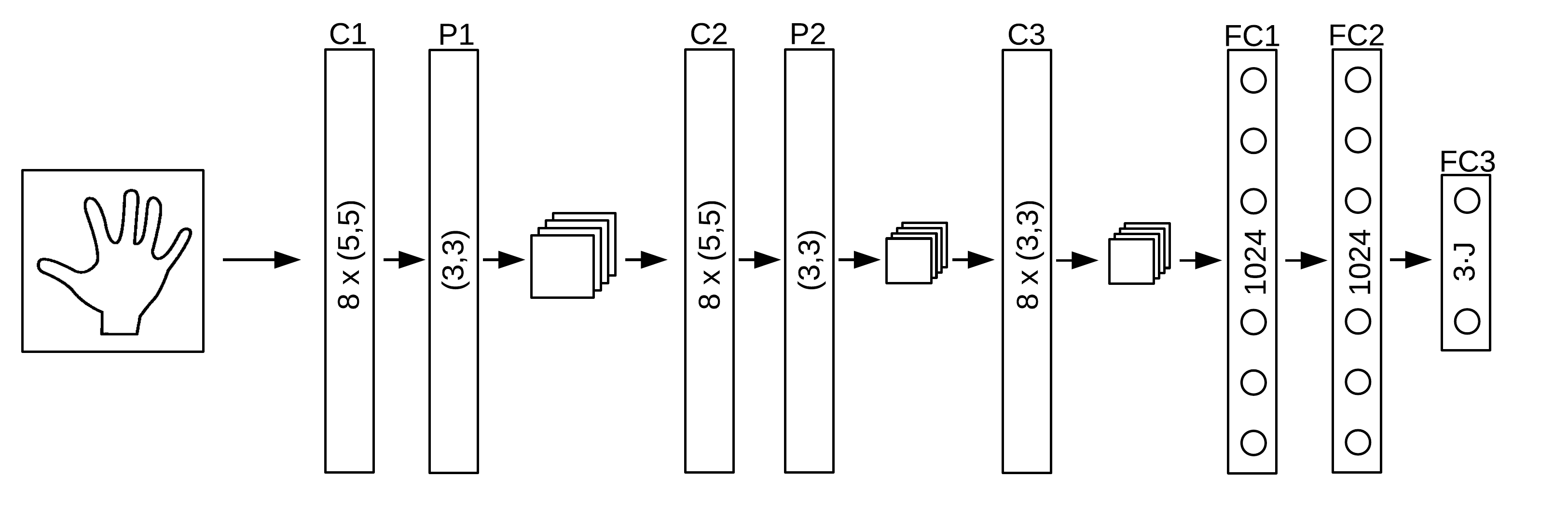}\label{fig:deepnet}
}\\
\subfloat[]{
   \includegraphics[width=0.37\textwidth]{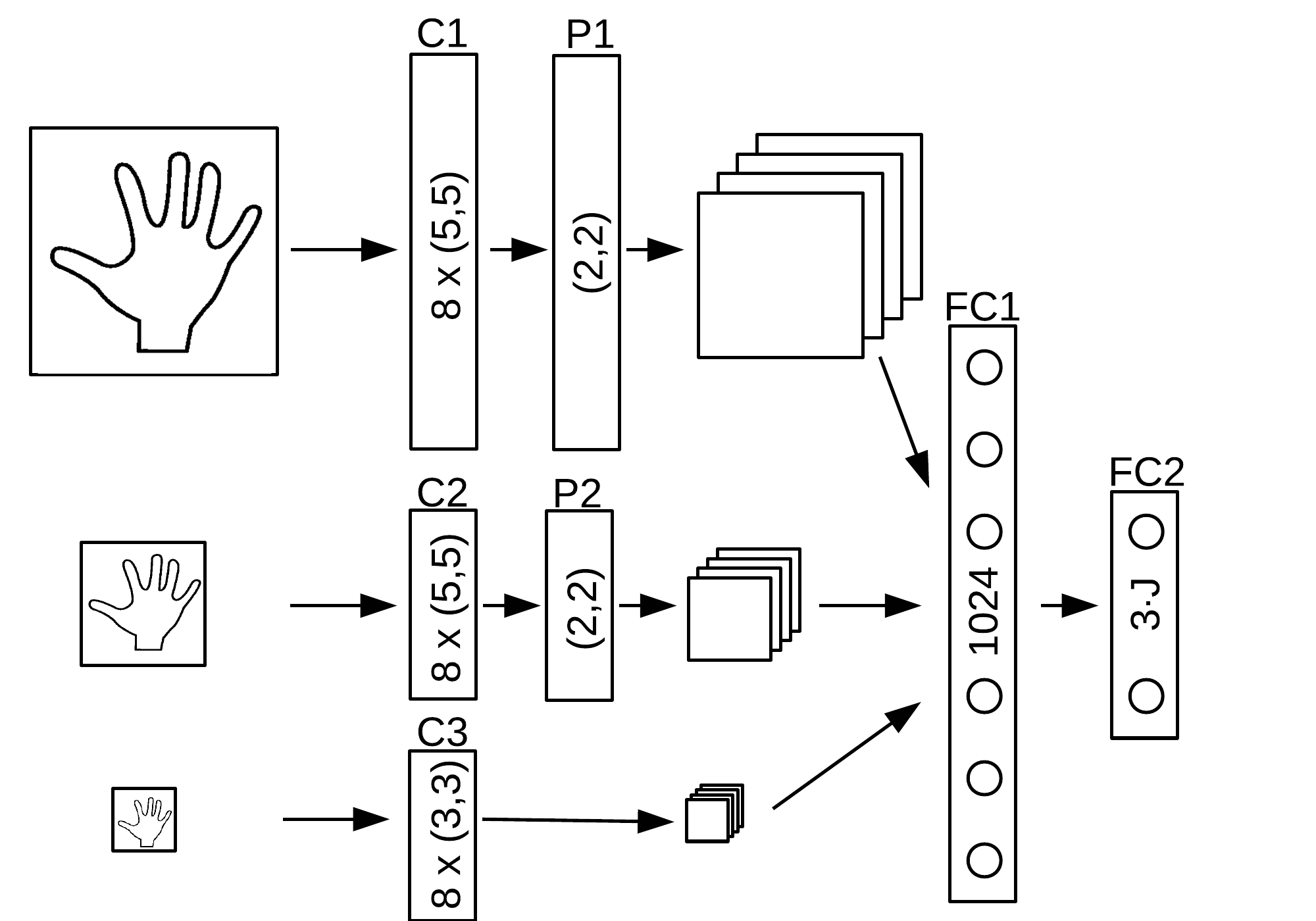}\label{fig:scalenet}
}\hspace{1cm}
\subfloat[]{
   \includegraphics[width=0.47\textwidth]{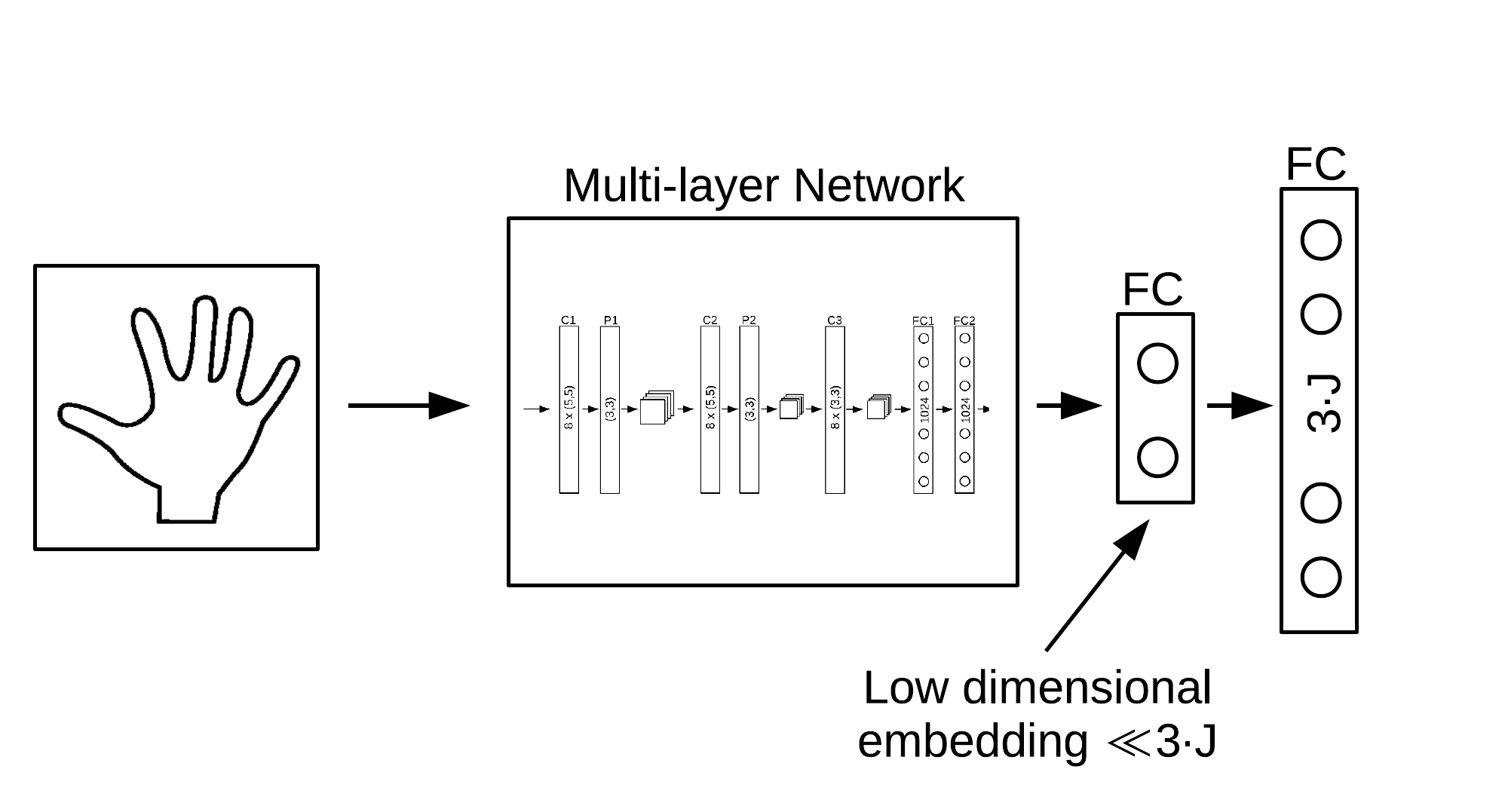}\label{fig:constrnet}
}
\end{center}
   \caption{\label{fig:nets}Different network architectures for the first stage.
     \textsf{C} denotes a convolutional layer with the number of filters and the
     filter size inscribed, \textsf{FC} a  fully connected layer with the number
     of  neurons, and  \textsf{P} a  max-pooling layer  with the  pooling size.   We
     evaluated      the      performance      of     a      shallow      network
     \protect\subref{fig:shallownet}       and       a      deeper       network
     \protect\subref{fig:deepnet},  as   well  as  a   multi-scale  architecture
     \protect\subref{fig:scalenet},           which           was           used
     in~\cite{Farabet2013,Sermanet2014}.   This  architecture extracts  features
     after   downscaling    the   input   depth   map    by   several   factors.
     \protect\subref{fig:constrnet}  All  these  networks  can  be  extended  to
     incorporate the constrained pose prior.   This causes an unusual bottleneck
     with less neurons than the output layer.  }
\label{fig:architecture_nets}
\end{figure*}

\subsection{Network Structures for Predicting the Joints' 3D Locations}

We first considered  two standard CNN architectures.   The first one is  shown in
Fig.~\ref{fig:shallownet}, and is a simple  shallow network, which consists of a
single convolutional  layer, a max-pooling  layer, and a  single fully-connected
hidden   layer.    The  second   architecture   we   consider  is   shown   in
Fig.~\ref{fig:deepnet}    and     is    a     deeper    but     still    generic
network~\cite{Krizhevsky2012,Toshev2014},   with   three  convolutional   layers
followed by max-pooling layers and two fully-connected hidden layers. All layers use Rectified Linear Unit~\cite{Krizhevsky2012} activation functions.

Additionally, we   evaluated   a   multi-scale    approach,   as   done   for   example
in~\cite{Farabet2013,Sermanet2014,Tompson2014a}. The  motivation  for  this  approach  is  that  using
multiple scales may help  capturing contextual information.  It uses  several downscaled
versions  of   the  input  image   as  input  to   the  network,  as   shown  in
Fig.~\ref{fig:scalenet}.

Our results will show that, unsurprisingly, the multi-scale approach performs better than the
deep architecture, which performs better than the shallow one. However, our
contributions, described in the next two sections, bring significantly more improvement.

\subsection{Enforcing a Prior on the 3D Pose}

So far  we only considered predicting  the 3D positions of  the joints directly.
However,  given  the  physical  constraints  over the  hand,  there  are  strong
correlation   between   the  different   3D   joint   locations,  and   previous
work~\cite{Wu2001} has shown  that a low dimensional embedding  is sufficient to
parameterize  the hand's  3D pose.  Instead of  directly predicting  the 3D  joint
locations,  we can  therefore predict  the  parameters of  the pose  in a  lower
dimensional  space.  As  this enforces  constraints of  the hand  pose, it  can be
expected that it  improves the  reliability of the predictions, which will be
confirmed by our experiments.

As shown in Fig.~\ref{fig:constrnet}, we implement the pose prior into the network structure by introducing a ``bottleneck'' in the last layer. This bottleneck is a layer with less neurons than necessary for the full pose representation,~\ie $\ll 3\cdot J$. It forces the network to learn a  low dimensional representation  of the  training data, that implements the physical constraints of the hand. Similar to  \cite{Wu2001}, we  rely on  a linear embedding. The embedding is enforced by the bottleneck layer and  the reconstruction from  the embedding to pose space is integrated as  a separate hidden layer
added  on top  of  the  bottleneck layer.  The weights of the reconstruction layer are set to compute the back-projection into the $3\cdot J$-dimensional joint space.  The resulting
network therefore directly  computes the full  pose. We initialize the reconstruction weights with the major components from a Principal Component Analysis of the hand pose data and then train the  full network using back-propagation. Using this approach we train the
networks described  in the previous  section. 

The embedding can be  as small as 8 dimensions for  a 42-dimensional pose vector
to fully represent the 3D pose as we show in the experiments.

\begin{figure*}
\begin{center}
\subfloat[]{
   \includegraphics[width=0.4\textwidth]{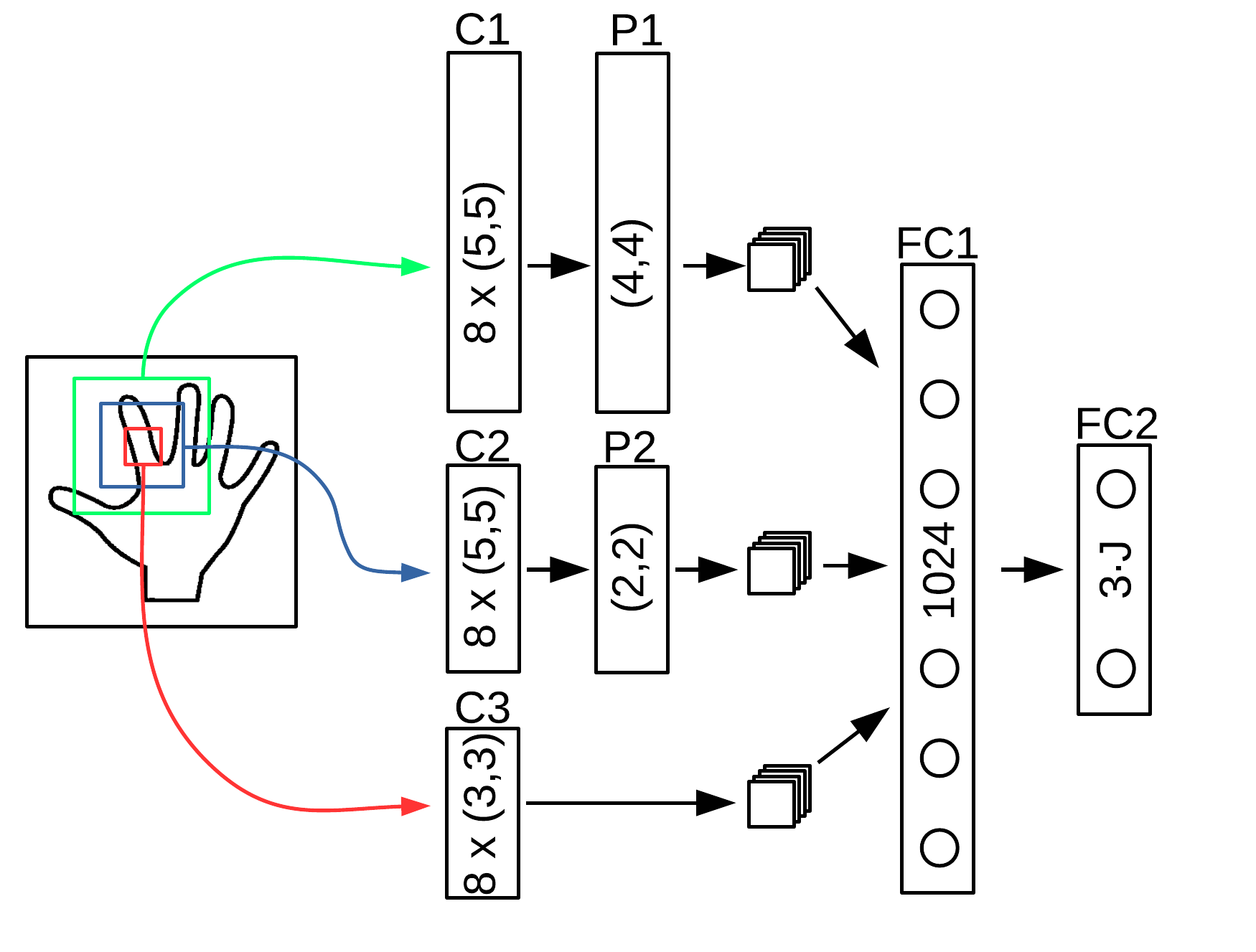}\label{fig:overlapnet}
}\hspace{1cm}
\subfloat[]{
   \includegraphics[width=0.51\textwidth]{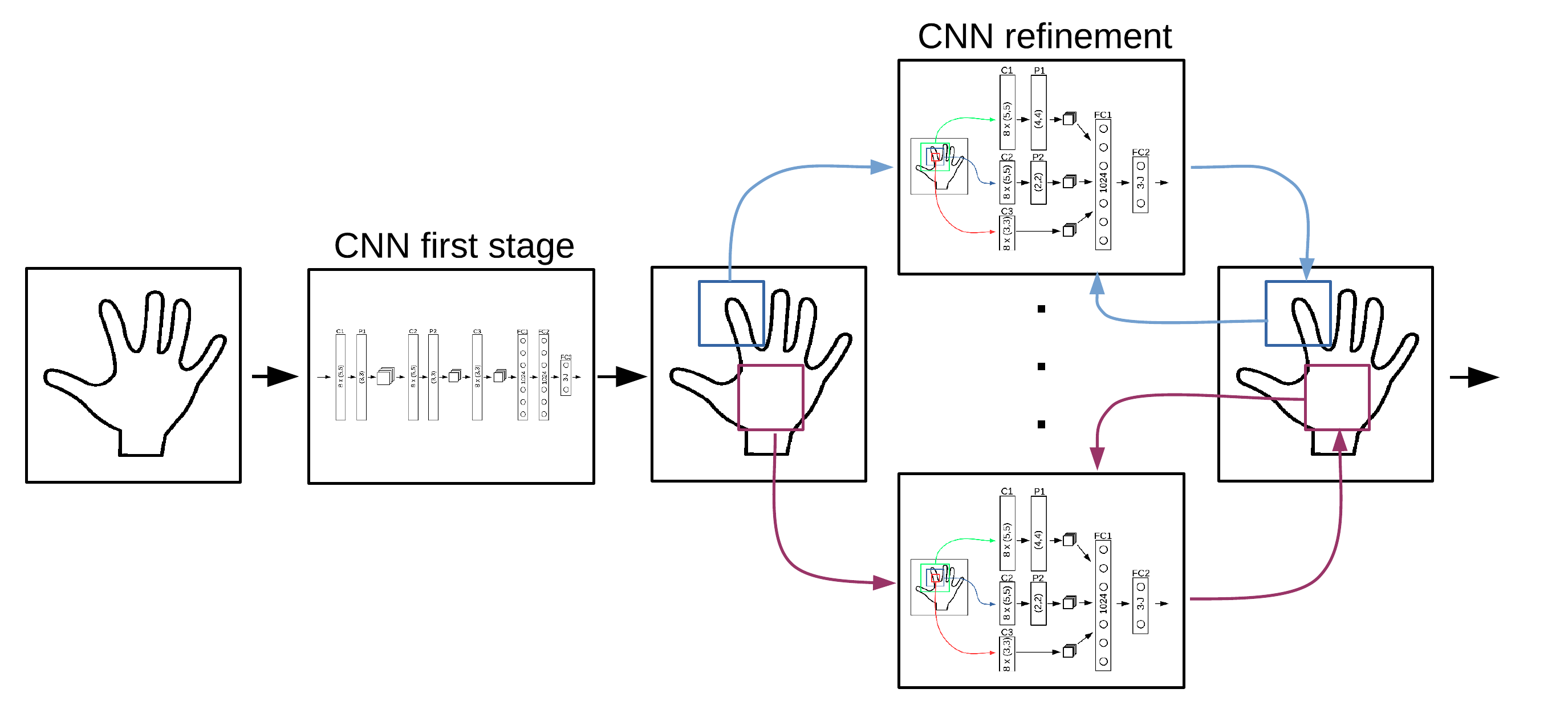}\label{fig:cascade}
}
\end{center}
\caption{Our architecture  for refining  the joint  locations during  the second
  stage.  We use a different network for each joint, to refine its location
  estimate as provided  by   the   first   stage.    \protect\subref{fig:overlapnet}   The
  architecture  we propose  uses overlapping  inputs  centered on  the joint  to
  refine.  Pooling  with small regions is  applied to the smaller  inputs, while
  the larger  inputs are pooled  with larger  regions. The smaller  inputs allow for higher
  accuracy, the  larger ones provide contextual  information.  We experimentally
  show that  this architecture  is more  accurate than  a more  standard network
  architecture.  \protect\subref{fig:cascade} shows a generic architecture of an
  iterative refinement,  where the output of  the previous iteration is  used as
  input  for  the  next.  As   for  Fig.~\ref{fig:nets},  \textsf{C}  denotes  a
  convolutional layer,  \textsf{FC} a  fully connected  layer, and  \textsf{P} a
  max-pooling layer. (Best viewed in color)}
\label{fig:architecture_casc}
\end{figure*}

\subsection{Refining the Joint Location Estimates}

The previous  architectures provide estimates for the   locations of
all the joints simultaneously. As done in~\cite{Sun2013,Toshev2014},
these estimates can then be refined independently.

Spatial context is important for this refinement step to avoid confusion between
the different fingers.  The best performing architecture we experimented with is
shown  in Fig.~\ref{fig:overlapnet}.   We  will refer  to  this architecture  as
\textit{ORRef},  for Refinement  with  Overlapping Regions.   It  uses as  input
several  patches of  different  sizes but  all centered  on  the joint  location
predicted by the first stage.  No pooling  is applied to the smallest patch, and
the size of the pooling regions then  increases with the size of the patch.  The
larger patches provide  more spatial context, whereas the absence  of pooling on
the small patch enables better accuracy.

We also considered  a standard CNN architecture as  a baseline, represented
in Fig.~\ref{fig:deepnet},  which relies on  a single input patch.   We will
refer  to  this  baseline  as  \textit{StdRef},  for  Refinement  with  Standard
Architecture.

To  further improve  the accuracy  of the  location estimates,  we iterate  this
refinement  step  several  times,  by  centering the  network  on  the  location
predicted at the previous iteration.


\section{Evaluation}
\label{sec:eval}

In  this section  we  evaluate  the different  architectures  introduced in  the
previous section  on several challenging  benchmarks.  We first  introduce these
benchmarks and  the  parameters  of our  methods.   Then  we describe  the
evaluation metric, and finally we present the results, quantitatively as well as
qualitatively. Our  results show that our  different contributions significantly
outperform the state-of-the-art.

\subsection{Benchmarks}

We evaluated our methods on the two following datasets:

\paragraph{NYU Hand Pose Dataset~\cite{Tompson2014}:}

This  dataset contains  over  72k training  and  8k test  frames  of RGB-D  data
captured  using the  Primesense Carmine  1.09.  It  is a  structured light-based
sensor  and the  depth  maps  have missing  values  mostly  along the  occluding
boundaries as well as noisy outlines.  For our experiments we use only the depth
data.  The dataset  has accurate annotations and exhibits a  high variability of
different poses.  The  training set contains samples from a  single user and the
test set samples from two different users.  The ground truth annotations contain
$J=36$ joints,  however \cite{Tompson2014} uses  only $J=14$ joints, and  we did
the same for comparison purposes.

\paragraph{ICVL Hand Posture Dataset~\cite{Tang2014}:} 

This dataset comprises a training set  of over 180k depth images showing various
hand poses.   The test set  contains two  sequences with each  approximately 700
depth  maps.  The  dataset is  recorded  using a  time-of-flight Intel  Creative
Interactive  Gesture  Camera and  has  $J=16$  annotated joints.   Although  the
authors provide different artificially rotated training samples, we only use the
genuine 22k.  The depth images have a high quality with hardly any missing depth
values, and sharp  outlines with little noise. However, the  pose variability is
limited compared to the NYU dataset.  Also, a relatively large number of samples
both from  the training and test  sets are incorrectly annotated: We evaluated the accuracy and about 36\% of
the poses from the test set have an annotation error of at least 10~mm.

\subsection{Meta-Parameters and Optimization}

The performance  of neural networks  depends on several meta-parameters,  and we
performed  a large  number of  experiments varying  the meta-parameters  for the
different architectures  we evaluated.  We report  here only the results  of the
best  performing sets  of  meta-parameters  for each  method.   However, in  our
experiments, the performance depends more on the architecture itself than on the
values of the meta-parameters.

We  trained the  different architectures  by  minimizing the  distance
between the  prediction and the expected  output per joint, and a regularization  term for
weight  decay to prevent over-fitting,  where  the  regularization   factor  is  $0.001$. We do not differ between occluded and non-occluded joints. Because  the
annotations are noisy, we use the robust Huber loss~\cite{Huber1964} to evaluate
the differences. The networks are trained with back-propagation using Stochastic
Gradient Descent~\cite{Bottou2010} with a batch  size of $128$ for $100$ epochs.
The learning rate is set to $0.01$ and we use a momentum of $0.9$~\cite{Polyak1964}.

\subsection{Evaluation Metrics}

We use two different evaluation metrics:
\begin{itemize}
\item the average Euclidean distance between the predicted 3D joint location and
  the ground truth, and
\item the fraction of test samples that have all predicted joints below a given
  maximum Euclidean distance from the ground truth, as was done
  in~\cite{Taylor2012}. This metric is generally regarded very challenging, as a
  single dislocated joint deteriorates the whole hand pose.
\end{itemize}

\begin{figure*}[t]
\begin{center}
\subfloat[Pose Prior on NYU dataset]{
   \includegraphics[width=0.47\linewidth]{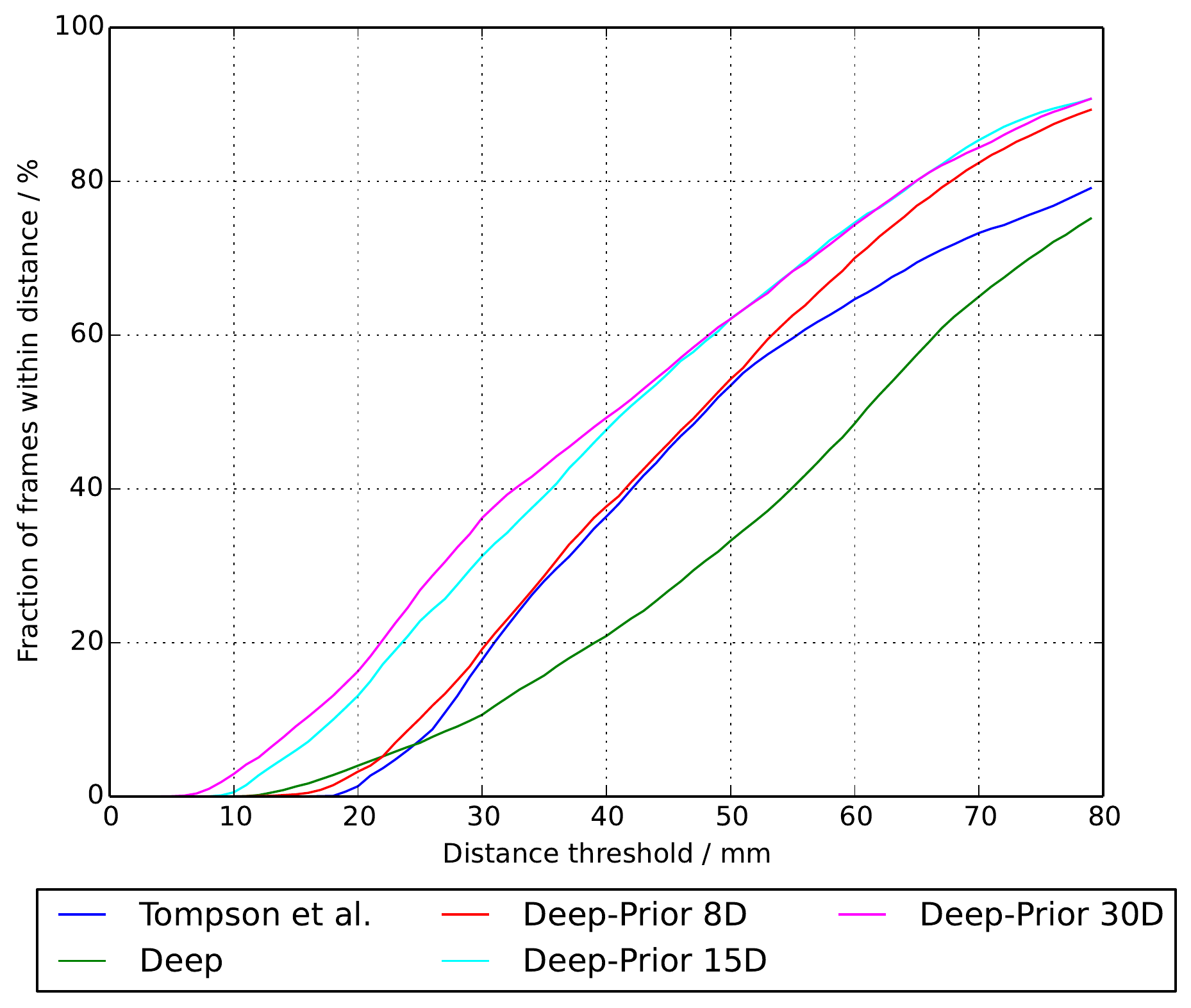}\label{fig:eval_frameswithin_NYU_DIMRED}
}
\subfloat[Refinement on NYU dataset]{
   \includegraphics[width=0.47\linewidth]{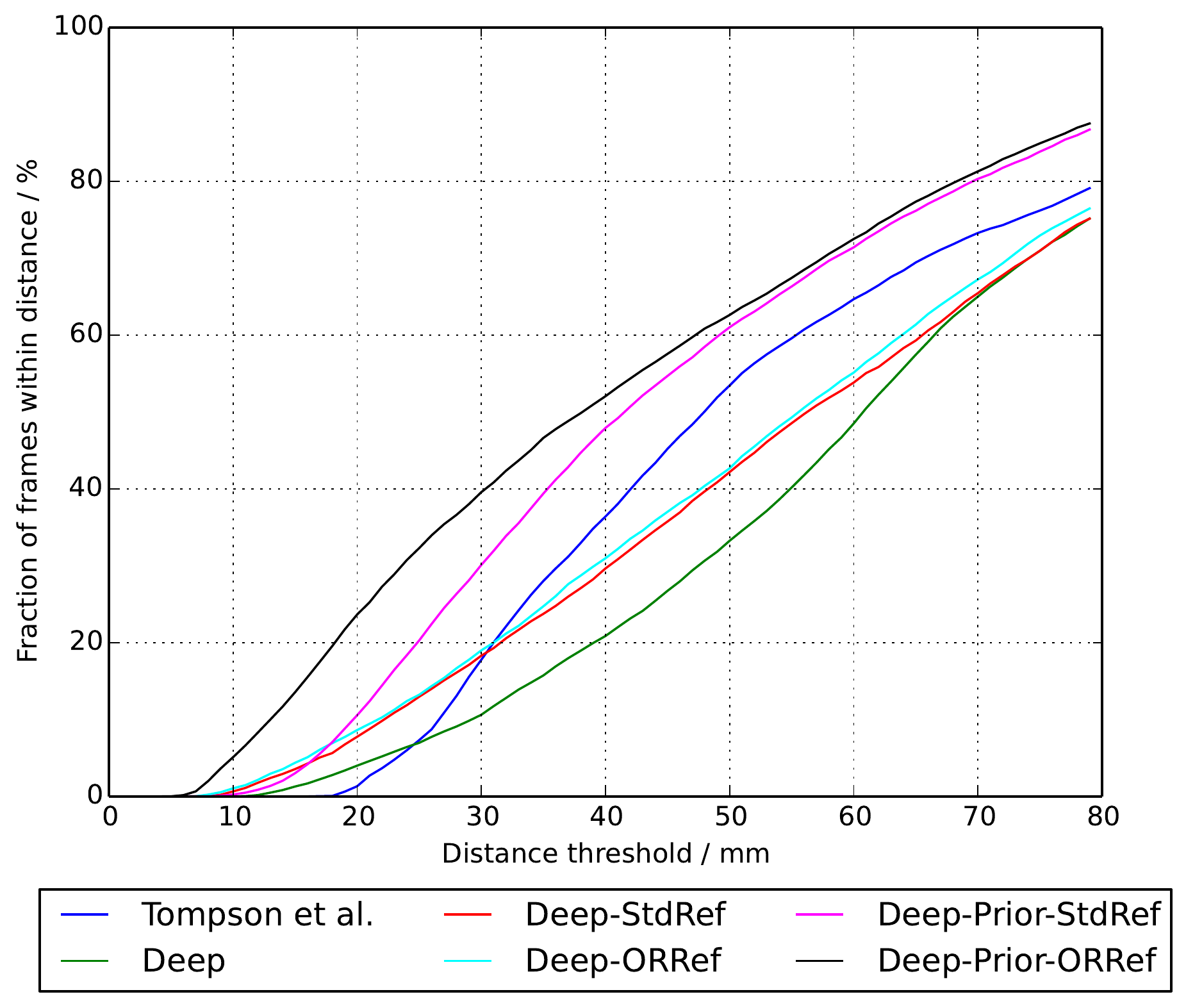}\label{fig:eval_frameswithin_NYU_REF}
}
\\
\subfloat[Pose Prior on ICVL dataset]{
   \includegraphics[width=0.47\linewidth]{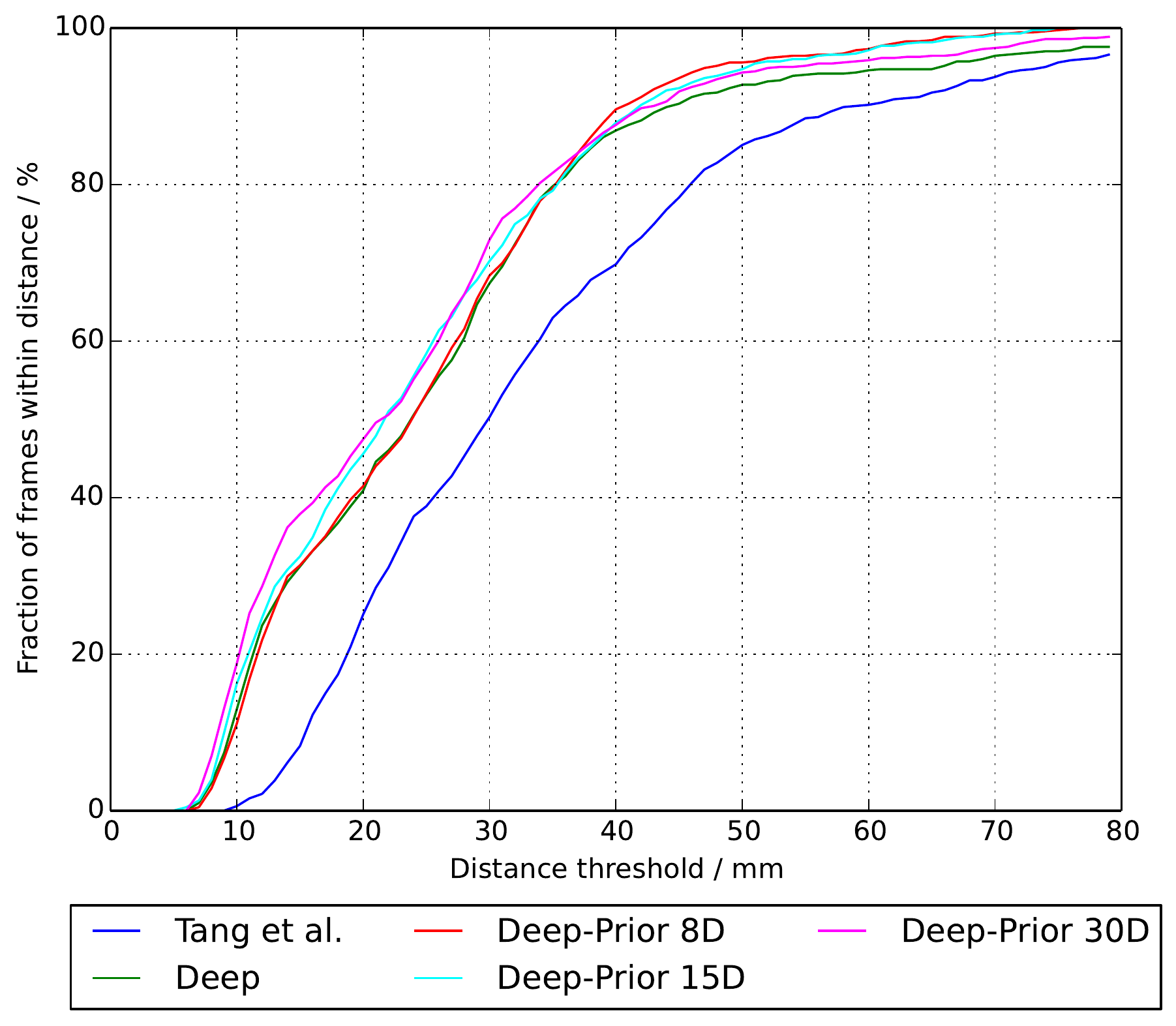}\label{fig:eval_frameswithin_ICVL_DIMRED}
}
\subfloat[Refinement on ICVL dataset]{
   \includegraphics[width=0.47\linewidth]{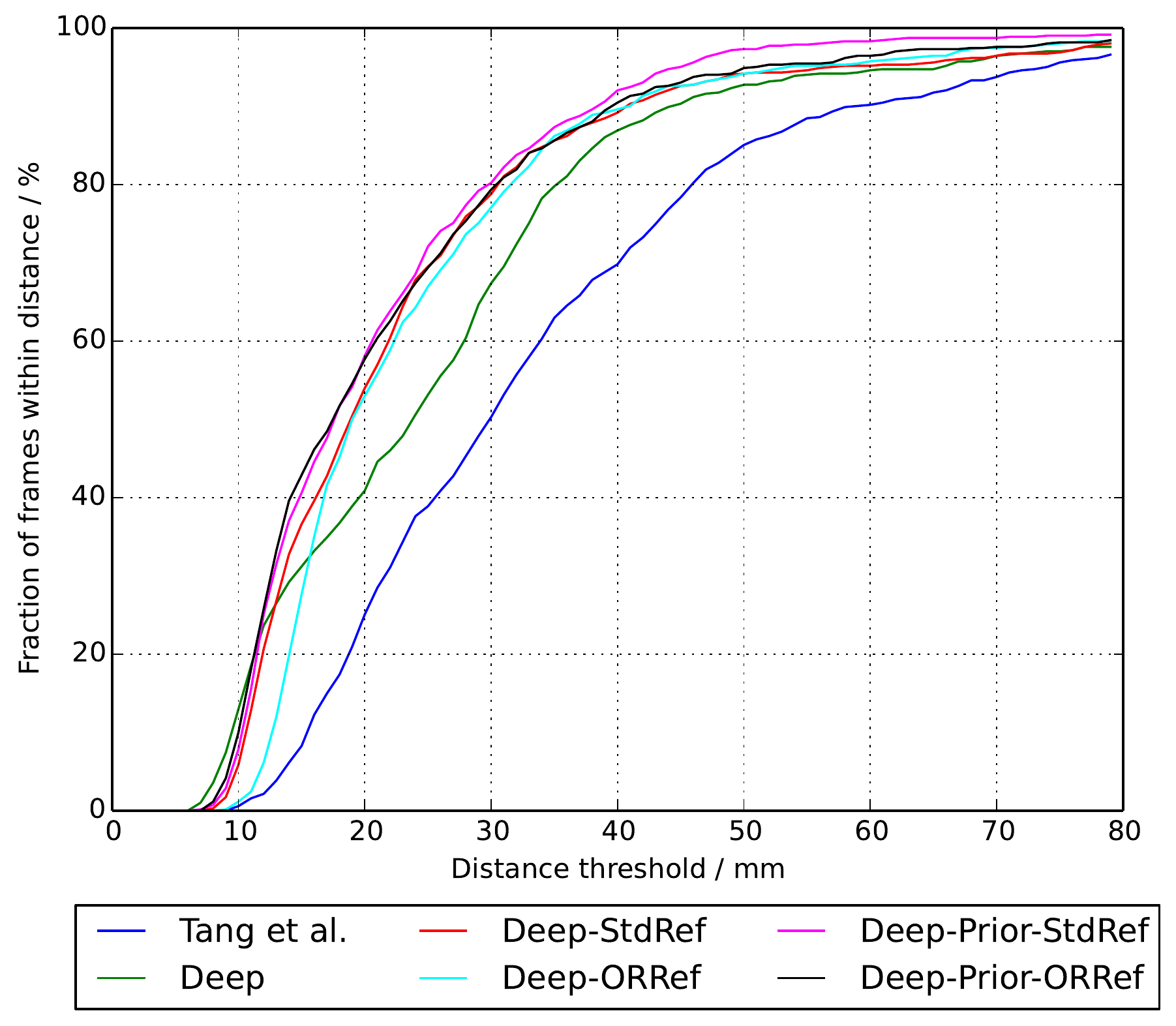}\label{fig:eval_frameswithin_ICVL_REF}
}
\end{center}
   \caption{Importance  of  the  pose  prior (left)  and  the  refinement  stage
     (right). We evaluate  the fraction of frames where all  joints are within a
     maximum distance for  different approaches.  A higher area  under the curve
     denotes                more                accurate                results.
     \textbf{Left \protect\subref{fig:eval_frameswithin_NYU_DIMRED}, 
     \protect\subref{fig:eval_frameswithin_ICVL_DIMRED}:}  We show  the influence
     of the dimensionality  of the pose embedding.  The optimal  value is around
     30, but using only 8 dimensions performs already very well.  The pose prior
     allows us to significantly outperform the state-of-the-art, even before the
     refinement   step. \textbf{Right \protect\subref{fig:eval_frameswithin_NYU_REF},
     \protect\subref{fig:eval_frameswithin_ICVL_REF}:}    We   show    that   our
     architecture with overlapping input  patches, denoted by the \textit{ORRef}
     suffix, provides higher accuracy for  refining the joint positions compared
     to a  standard deep CNN,  denoted by  the \textit{StdRef} suffix.   For the
     baseline of  Tompson~\etal~\cite{Tompson2014} we  augment their 2D joint locations with the  depth
     from the depth maps, as done by~\cite{Tompson2014}, and depth values that  do not lie within
     the hand  cube are truncated  to the cube's back  face to avoid  large
     errors. (Best viewed on screen)}
\label{fig:results_quantitative_pose}
\end{figure*}

\subsection{Importance of the Pose Prior}

In                                   Fig.~\ref{fig:eval_frameswithin_NYU_DIMRED}
and~\ref{fig:eval_frameswithin_ICVL_DIMRED}   we  compare   different  embedding
dimensions and direct regression in the full $3\cdot J$-dimensional pose space for  the NYU and the  ICVL dataset, respectively.  The  evaluation on
both  datasets  shows  that enforcing a pose prior is beneficial compared to direct regression in the full pose space. Only  8  dimensions out  of  the  original  42-  or
48-dimensional pose spaces are already enough to capture the pose and outperform
the baseline on  both datasets.  However, the  30-dimensional embedding performs
best, and thus we  use this for all further evaluations.  The  results on the ICVL
dataset, which has  noisy annotations, are not as drastic,  but still consistent
with the results on the NYU dataset.

The baseline on the NYU dataset of Tompson~\etal~\cite{Tompson2014} only provide
the 2D  locations of the  joints. For comparison,  we follow their  protocol and
augment their 2D locations  by taking the depth of each  joint directly from the
depth maps  to derive  comparable 3D  locations.  Depth values  that do  not lie
within the hand cube  are truncated to the cube's back face  to avoid large
errors. This protocol, however, has a  certain influence on the error metric, as
evident in  Fig.~\ref{fig:eval_jointmean_NYU_REF}.  The augmentation  works well
for some joints, as apparent by the  average error. However, it is unlikely that
the augmented  depth is  correct for all  joints of the  hand,~\eg the  2D joint
location lies on the background or  is self-occluded, thus causing higher errors
for individual  joints. When  using the evaluation  metric of~\cite{Taylor2012},
where all joints have to be within a maximum distance, this outlier has a strong
influence, in contrast to the evaluation  of the average error, where an outlier
can  be  insignificant for  the  mean.  Thus  we  outperform the  baseline  more
significantly for the distance threshold than for the average error.

\subsection{Increasing Accuracy with Pose Refinement}

The refinement  stage can be used  to further increase the  location accuracy of
the predicted  joints.  We achieved the  highest accuracy by using  our CNN with
constrained  prior hand  model  as first  stage, and  then  applying the  second
iterative  refinement  stage  with  our   CNN  with  overlapping  input
patches, denoted \textit{ORRef}.

\begin{figure*}[t]
\begin{center}
\subfloat[NYU dataset]{
   \includegraphics[width=0.45\linewidth]{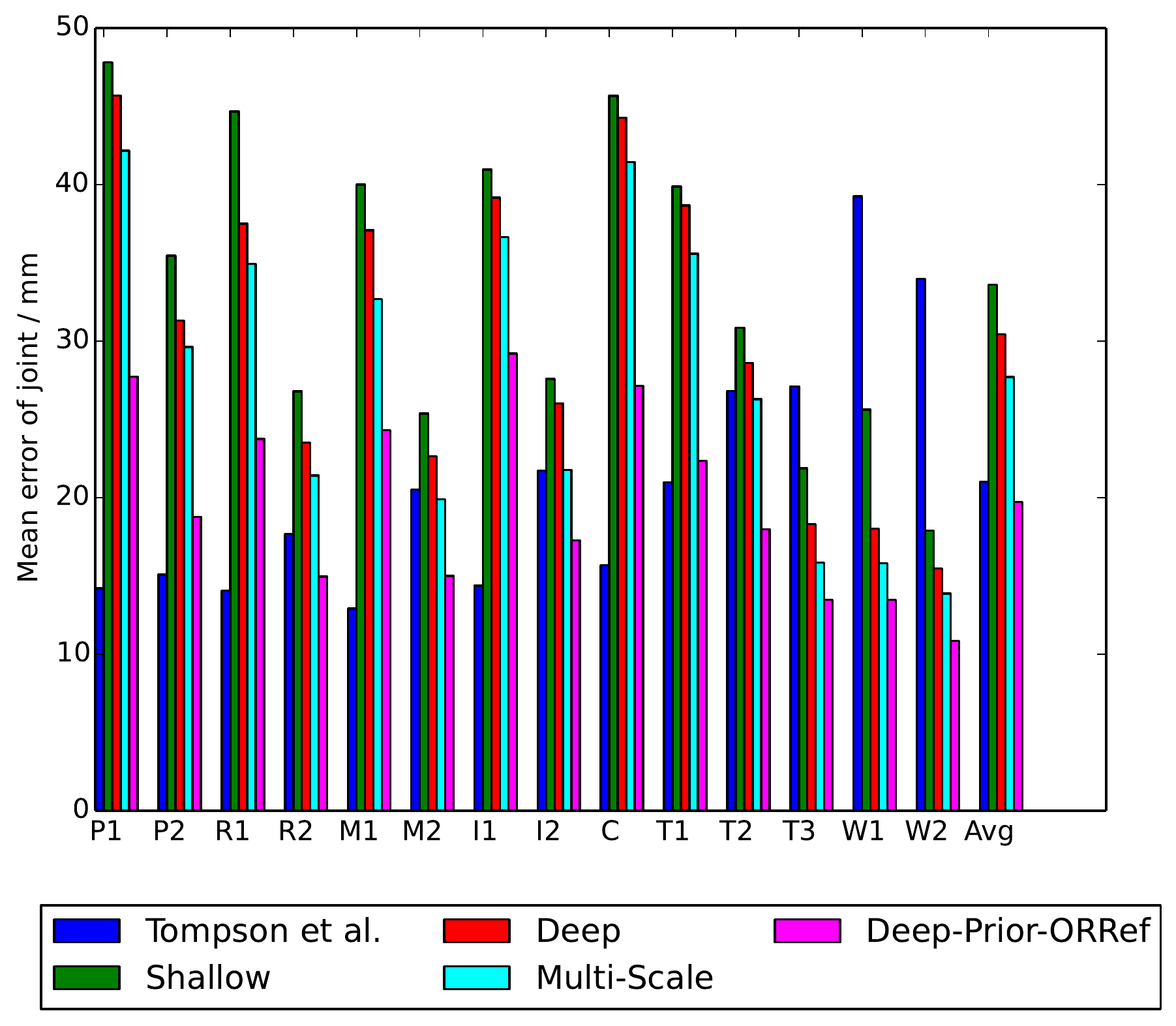}\label{fig:eval_jointmean_NYU_REF}
}
\subfloat[ICVL dataset]{
   \includegraphics[width=0.45\linewidth]{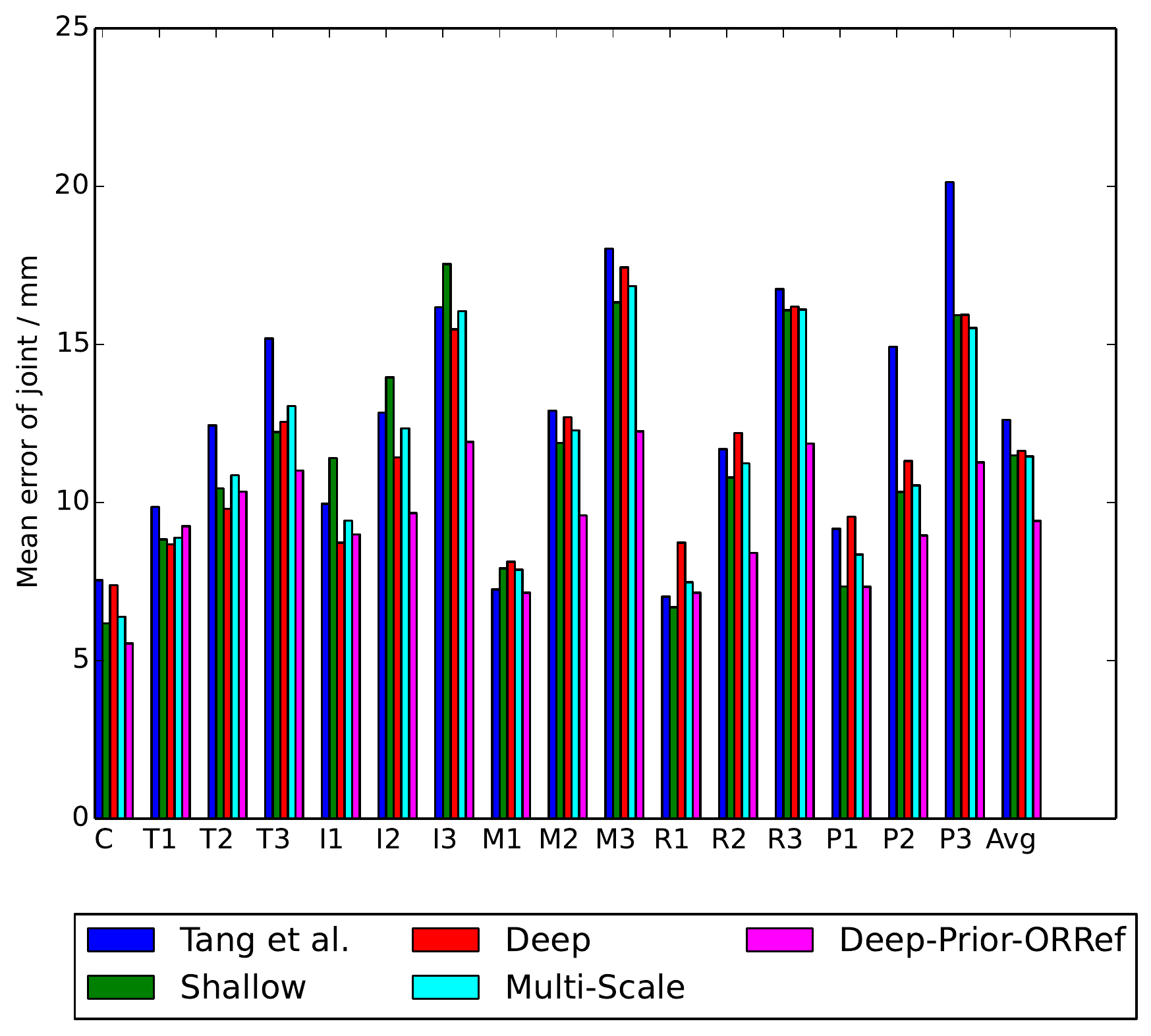}\label{fig:eval_jointmean_ICVL_REF}
}
\end{center}
   \caption{Average joint errors. For completeness and comparability we show the
     average joint errors, which are, however, not as decisive as the evaluation
     in  Fig.~\ref{fig:results_quantitative_pose}. Though, the   results  are
     consistent. The evaluation of the average error is more tolerant to larger errors of a single joint, which deteriorate the pose as for Fig.~\ref{fig:results_quantitative_pose}, but are insignificant for the mean if the other joints are accurate.      Our  proposed   architecture  \textit{Deep-Prior-ORRef},  the
     constrained pose CNN with refinement  stage, provides the highest accuracy.
     For the ICVL dataset, the  simple baseline architectures already outperform
     the baseline.  However,  they cannot capture the higher  variations in pose
     space and noisy  images of the NYU dataset, where  they perform much worse.
     The palm and fingers are indexed as C:  palm, T: thumb, I: index, M: middle,
     R: ring, P: pinky, W: wrist. (Best viewed on screen)}
\label{fig:results_quantitative_mean}
\end{figure*}

The         results         in         Fig.~\ref{fig:eval_frameswithin_NYU_REF},~\ref{fig:eval_frameswithin_ICVL_REF}                                    and~\ref{fig:results_quantitative_mean}  show  that   applying  the  refinement
improves the location  accuracy for different base CNNs.  From rather inaccurate initial
estimates, as  provided by the  standard deep  CNN, our proposed  ORRef performs
only   slightly   better   than   refinement  with   the   baseline   deep
CNN, denoted by \textit{StdRef}. This  is because for large  initial errors
only the  larger input  patch provides  enough context  for reasoning  about the
offset. The smaller input patch cannot  provide any information if the offset is
bigger than the patch size. For  more accurate initial estimates, as provided by
our deep  CNN with pose  prior, the ORRef takes  advantage from the  small input
patch which does not use pooling for higher accuracy. We iterate our  refinement two times, since iterating more  often does not provide
any further increase in accuracy.

We would like to emphasize that our results on the ICVL dataset, with an average
accuracy  below 10~mm,  already  scratch  at the  uncertainty  of the  labelled
annotations.  As already  mentioned, the  ICVL dataset  suffers from  inaccurate
annotations, as  we show  in some  qualitative samples in Fig.~\ref{fig:results_qualitative} first and fourth column.  While  this has  only a
minor effect  on training, the  evaluation is  more affected.  We  evaluated the
accuracy of  the test sequence  by revising the  annotations in image  space and
calculated an average error of 2.4~mm with a standard deviation of 5.2~mm.

\subsection{Running Times}

Table~\ref{tab:runtime} provides a comparison  of the running times of
the different methods,  both on CPU and GPU.  They  were measured on a
computer equipped  with an Intel  Core i7, 16GB  of RAM, and an nVidia
GeForce GTX  780 Ti GPU. Our  methods are implemented in  Python using
the  Theano library~\cite{Bergstra2010},  which  offers  an option  to
select  between the  CPU and  the  GPU for  evaluating CNNs.   Our
different models  perform very fast, up  to over 5000~fps on  a single
GPU.  Training  takes about five hours  for  each
CNN.  The deep  network with  pose  prior performs  very fast  and
outperforms all  other methods in  terms of accuracy. However,  we can
further refine the joint locations at the cost of higher runtime.

\begin{table}
\begin{center}
\begin{tabular}{@{}lll@{}}
\toprule
Architecture & GPU & CPU \\
\midrule
Shallow & 0.07~ms & 1.85~ms\\
Deep~\cite{Krizhevsky2012} & 0.1~ms & 2.08~ms\\
Multi-Scale~\cite{Farabet2013} & 0.81~ms & 5.36~ms\\
Deep-Prior & 0.09~ms & 2.29~ms\\
Refinement & 2.38~ms & 62.91~ms\\
Tompson~\etal~\cite{Tompson2014} & 5.6~ms & -\\
Tang~\etal~\cite{Tang2014} & - & 16~ms\\
\bottomrule
\end{tabular}
\end{center}
\caption{Comparison  of different  runtimes. Our  CNN with  pose prior
  (\textit{Deep-Prior}) is faster by a magnitude compared to the other
  methods (pose estimation only). We can further increase the accuracy
  using the refinement  stage, still at competitive speed.  All of the
  denoted  baselines  use   state-of-the-art  hardware  comparable  to
  ours.}\label{tab:runtime}
\end{table}

\subsection{Qualitative Results}

We present  qualitative results in  Fig.~\ref{fig:results_qualitative}. The
typical problems of  structured light-based sensors, such as  missing depth, can
be problematic for accurate localization. However, only partially missing parts,
as shown in the third and  fourth columns for example, do not significantly deteriorate the
result. The location of  the joint is constrained by the  learned hand model. If
the missing  regions get too  large, as shown in  the fifth column,  the accuracy
gets worse.  However, because of the use of the pose subspace embedding, the  predicted joint locations  still preserve  the learned
hand topology.  The erroneous  annotations of the  ICVL dataset  deteriorate the
results, as our  predicted locations during the first stage are sometimes more
accurate than the ones obtained during the second stage: see for example the pinky
in the  first  or  fourth  column. 

\bgroup
\setlength{\tabcolsep}{1pt}
\begin{figure*}[t]
\begin{center}
\begin{tabular}{@{}lcccc|c||cccc|c@{}}
& \multicolumn{5}{c||}{NYU dataset} & \multicolumn{5}{|c}{ICVL dataset} \\

\parbox[t]{3mm}{\multirow{2}{*}{\rotatebox[origin=c]{90}{Deep-Prior}}} &
\includegraphics[width=0.1\linewidth,trim={5cm 3cm 5cm 3cm},clip]{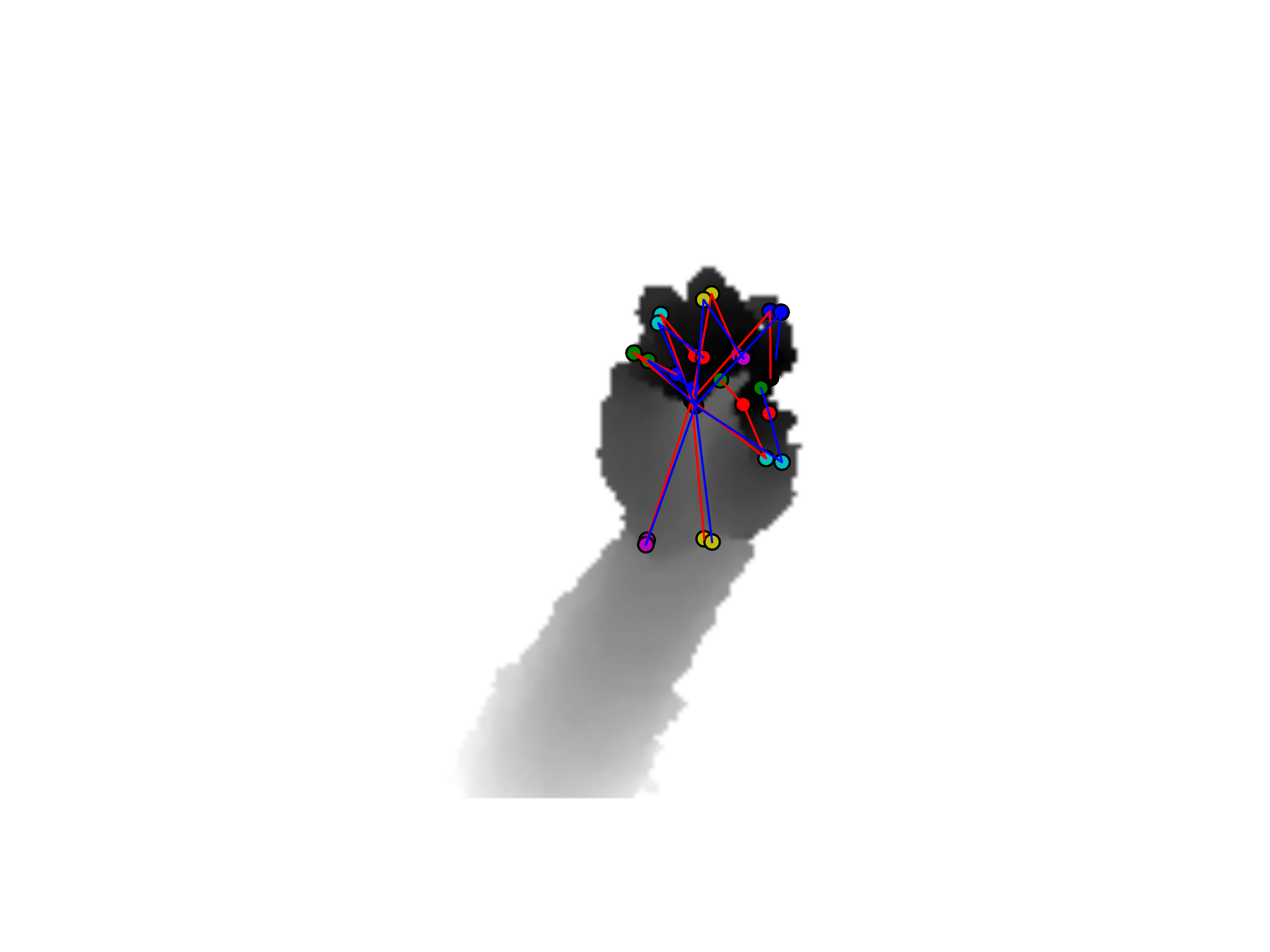} &
\includegraphics[width=0.1\linewidth,trim={5cm 3cm 5cm 3cm},clip]{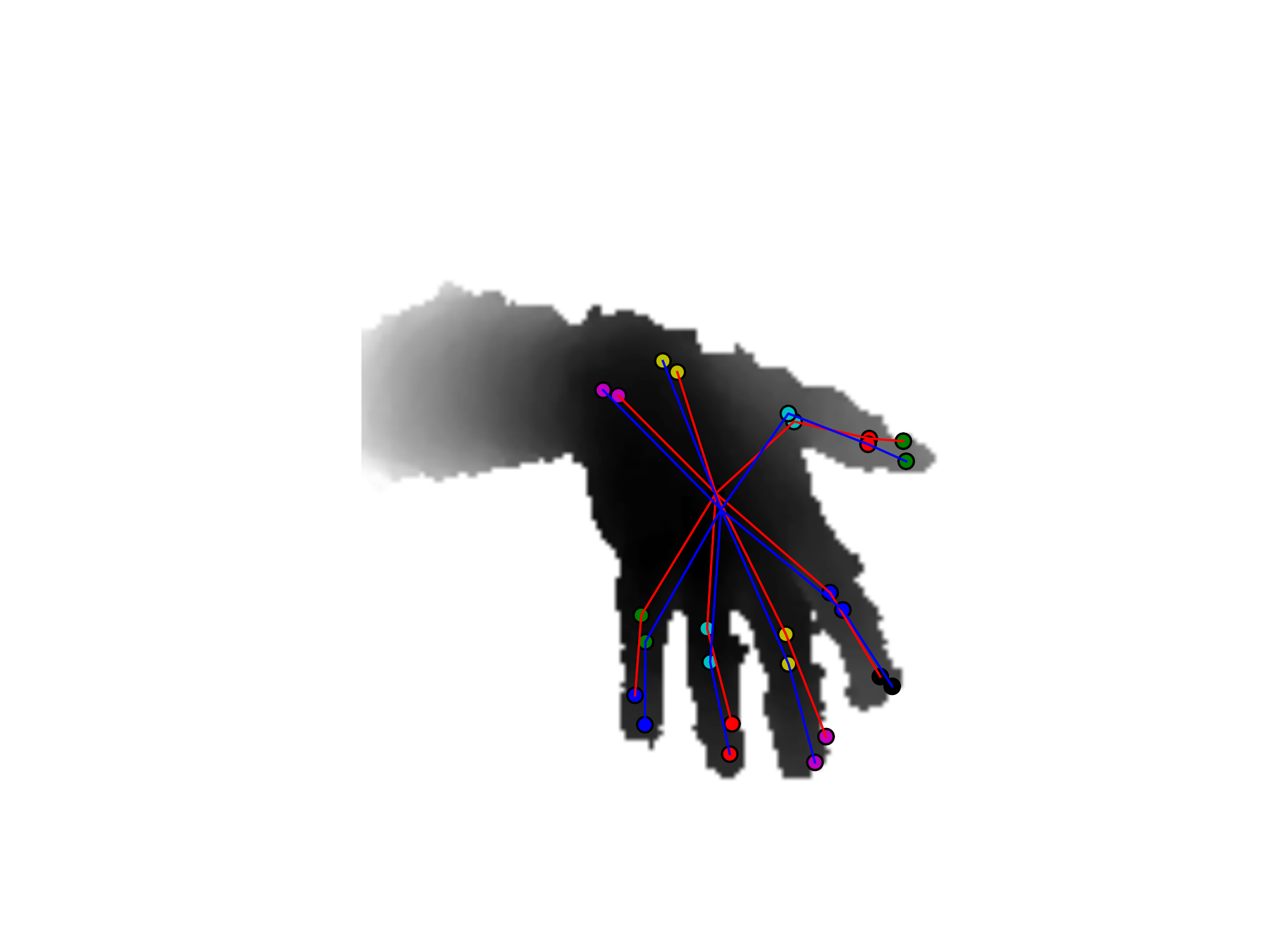} &
\includegraphics[width=0.1\linewidth,trim={5cm 3cm 5cm 3cm},clip]{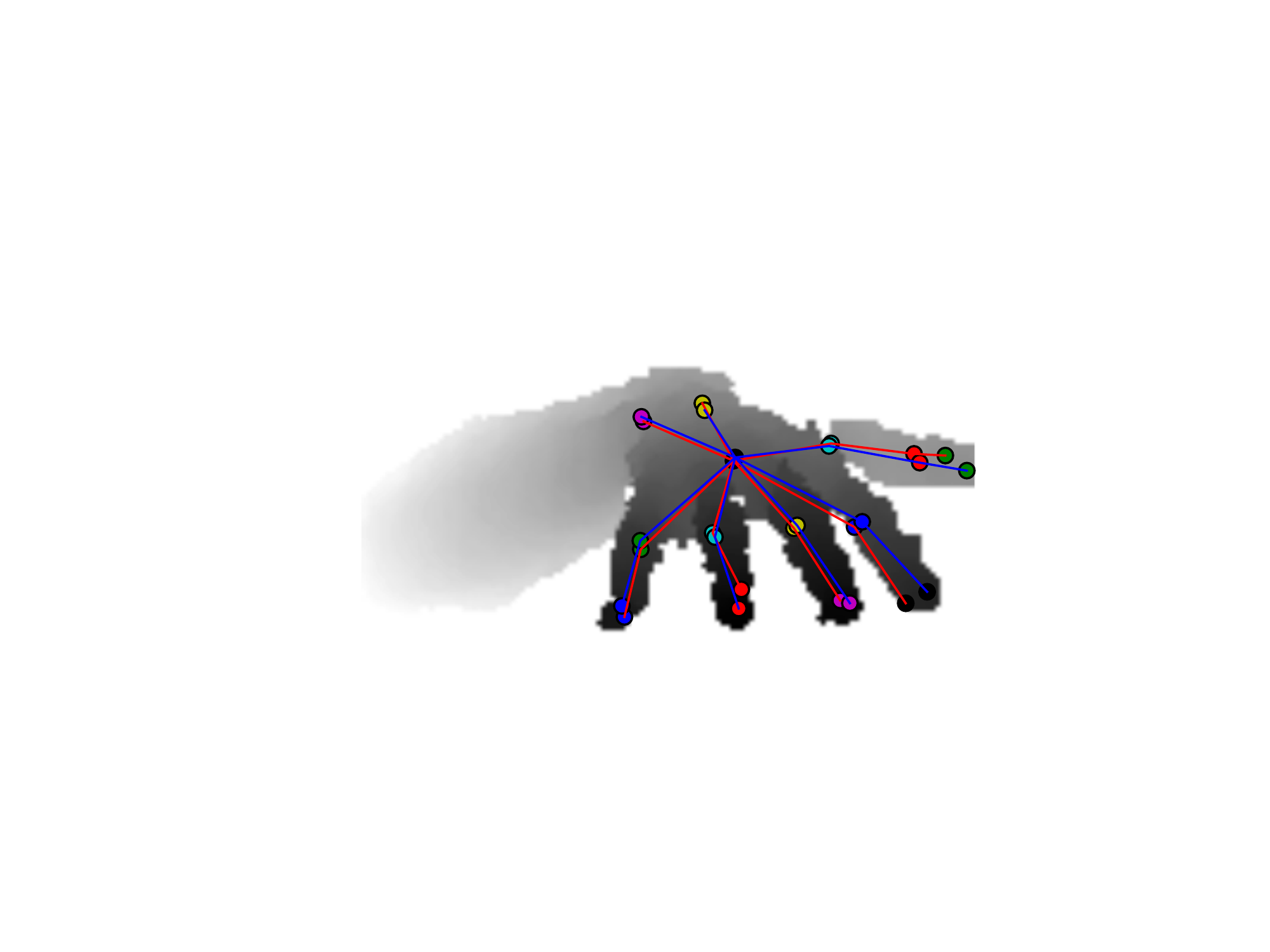} &
\includegraphics[width=0.1\linewidth,trim={5cm 3cm 5cm 3cm},clip]{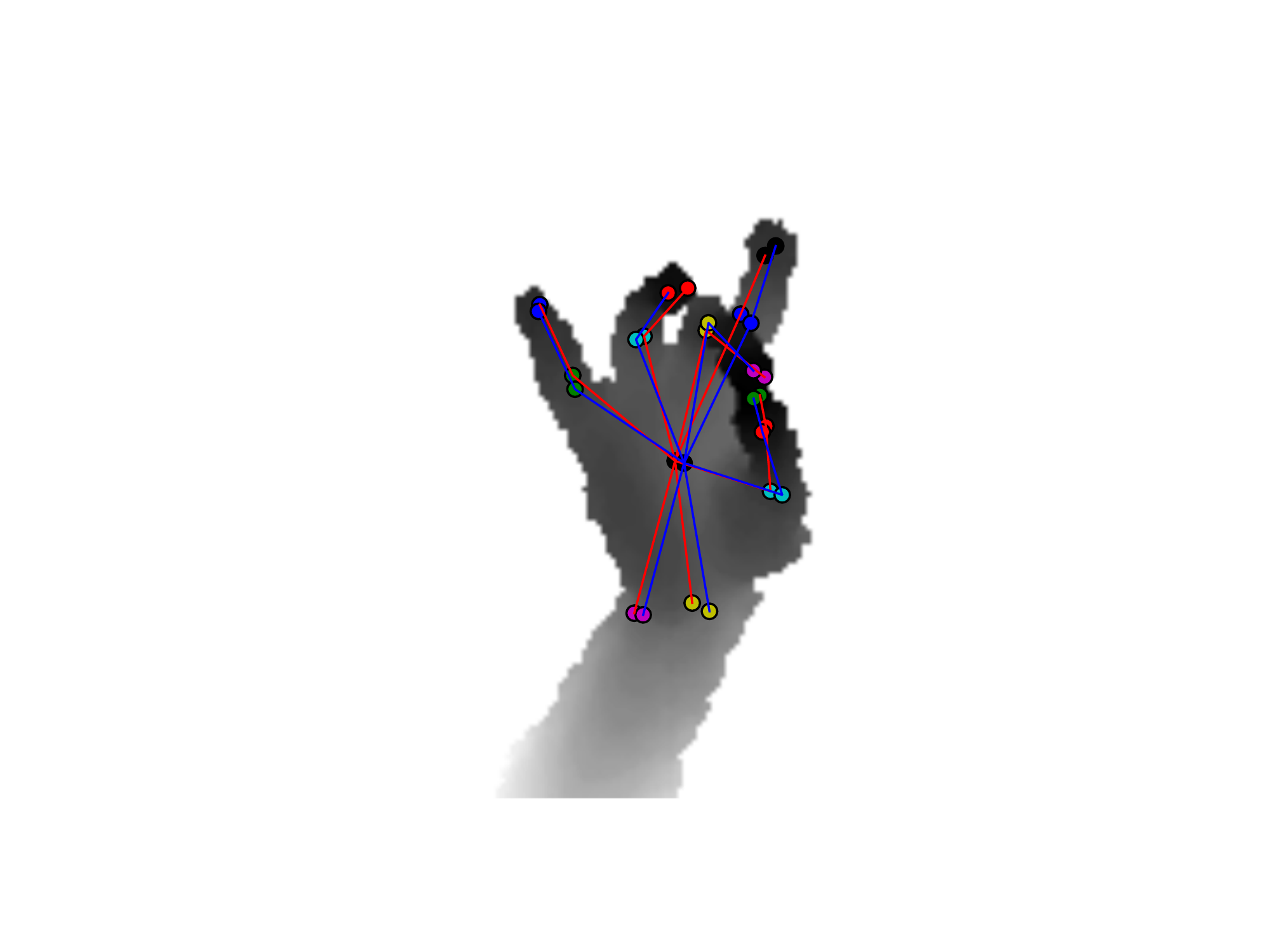} &

\includegraphics[width=0.1\linewidth,trim={5cm 3cm 5cm 3cm},clip]{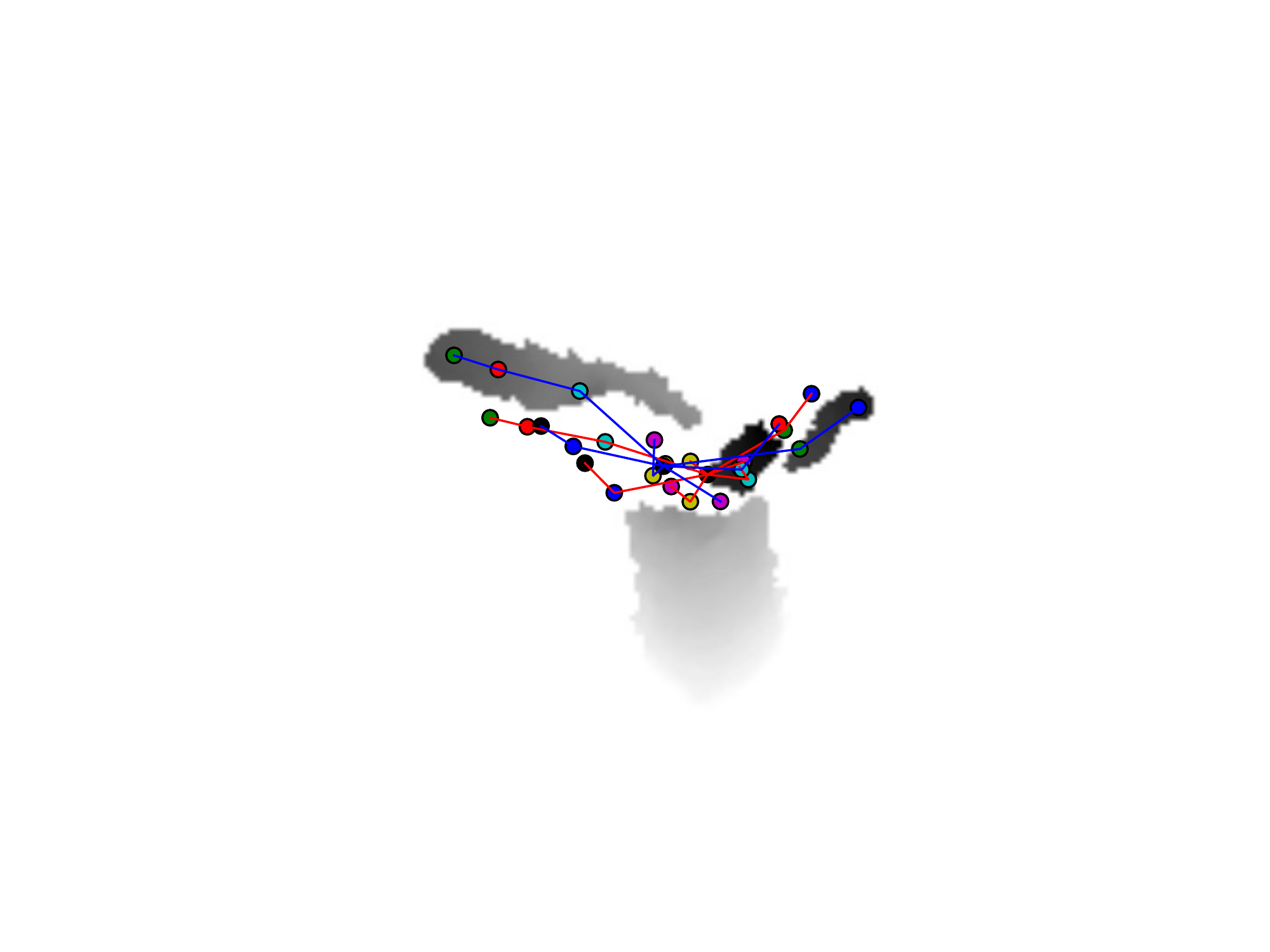} &

\includegraphics[width=0.1\linewidth,trim={5cm 3cm 5cm 3cm},clip]{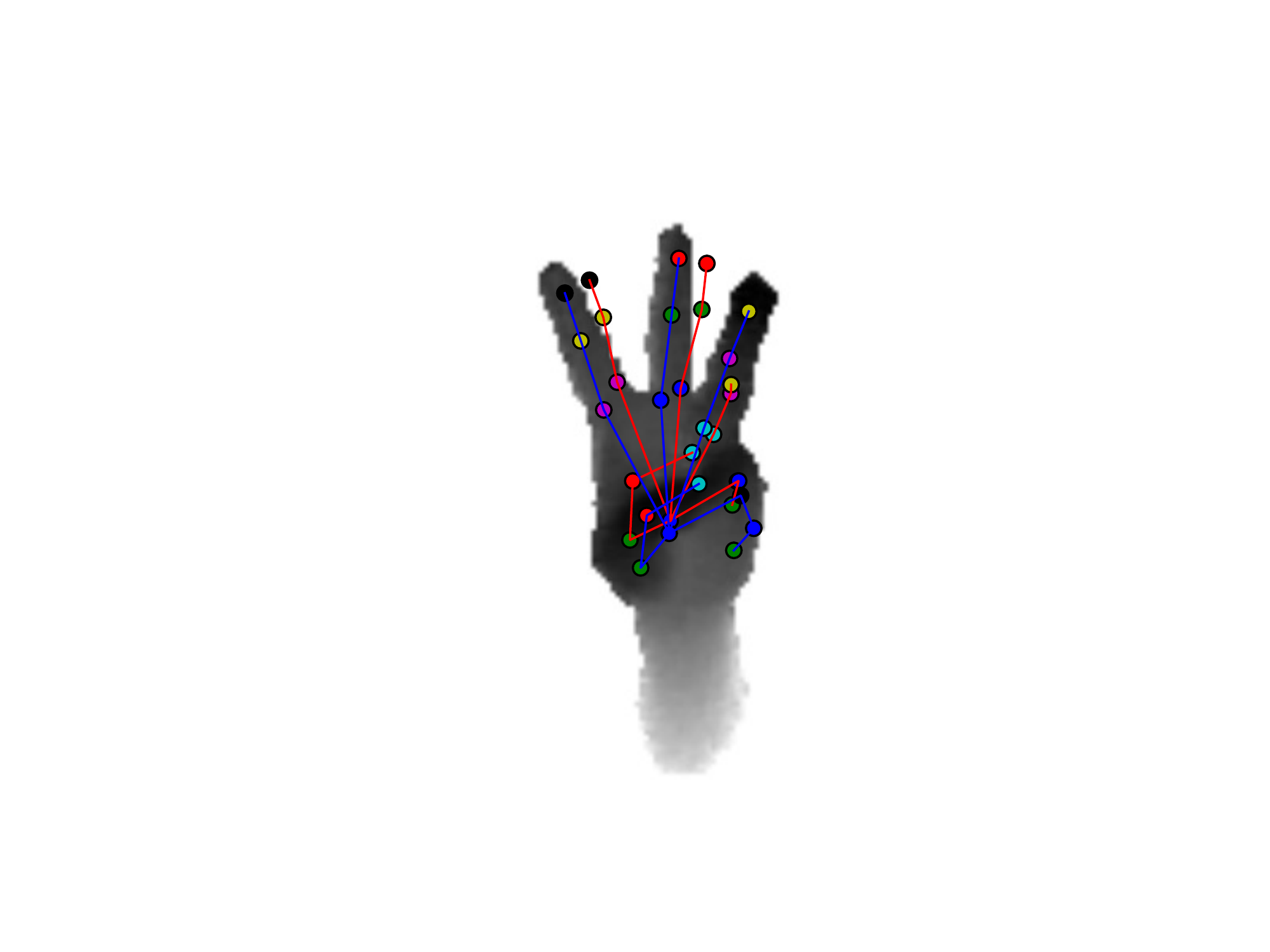} &
\includegraphics[width=0.1\linewidth,trim={5cm 3cm 5cm 3cm},clip]{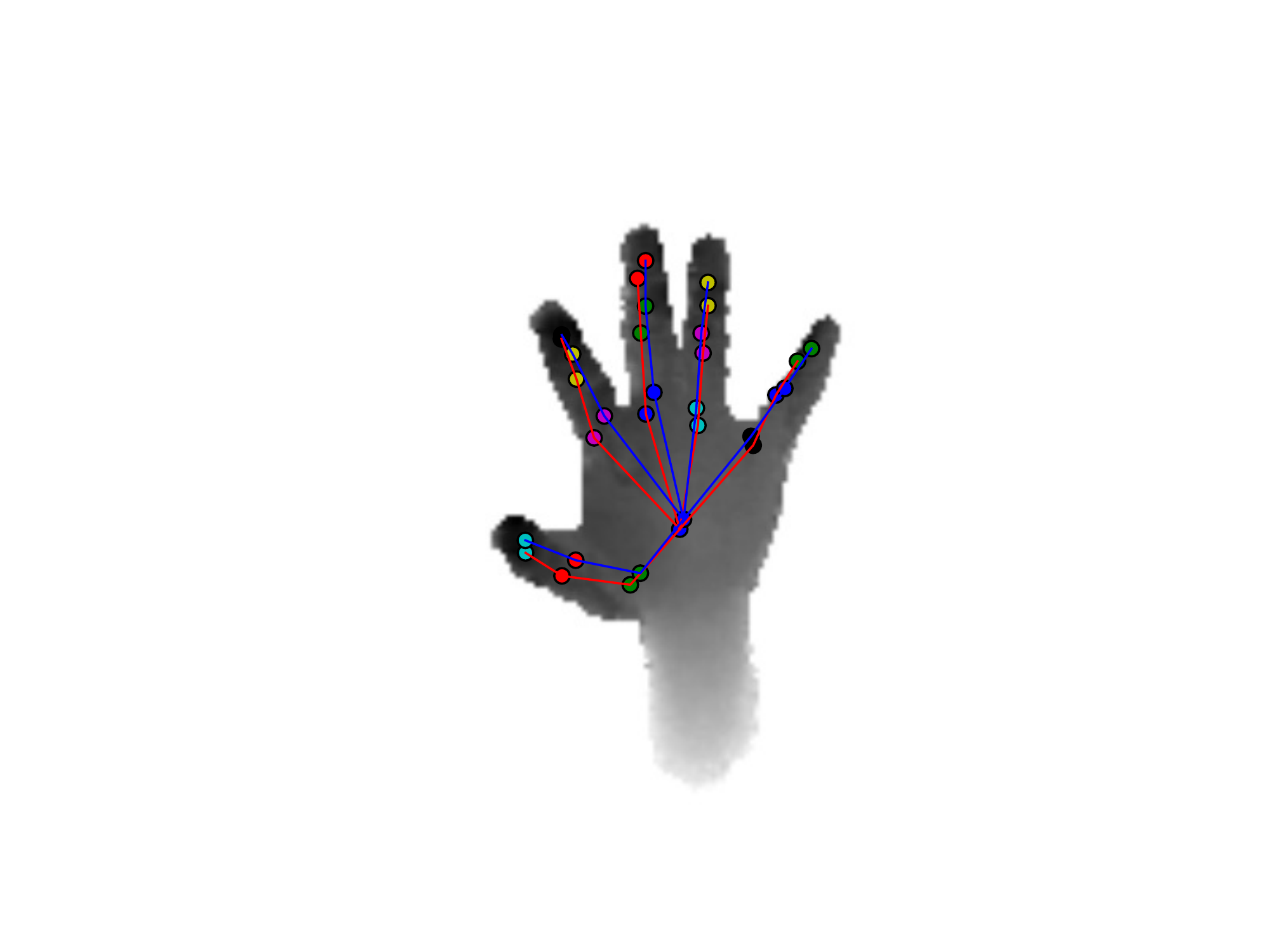} &
\includegraphics[width=0.1\linewidth,trim={5cm 3cm 5cm 3cm},clip]{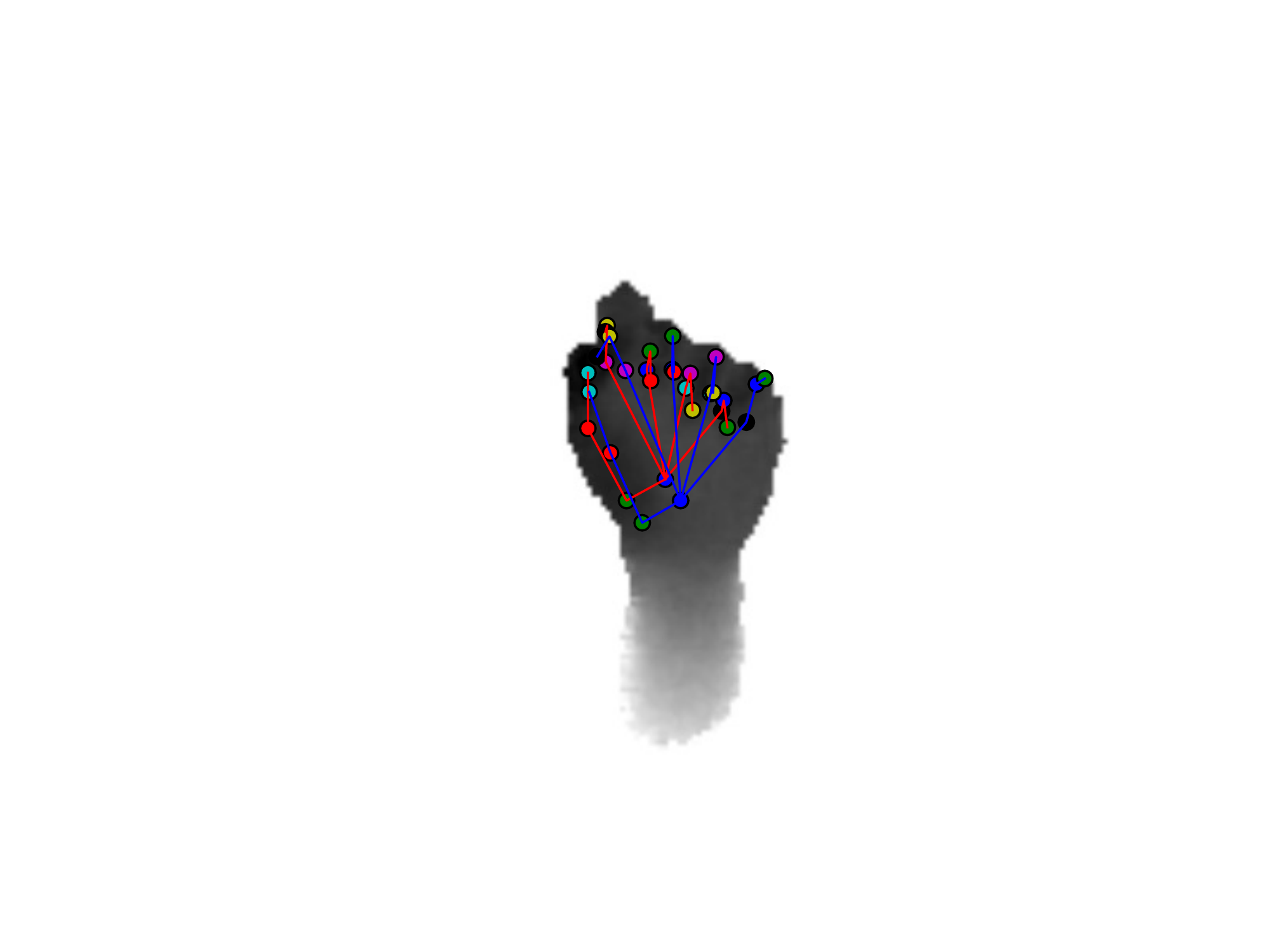} &
\includegraphics[width=0.1\linewidth,trim={5cm 3cm 5cm 3cm},clip]{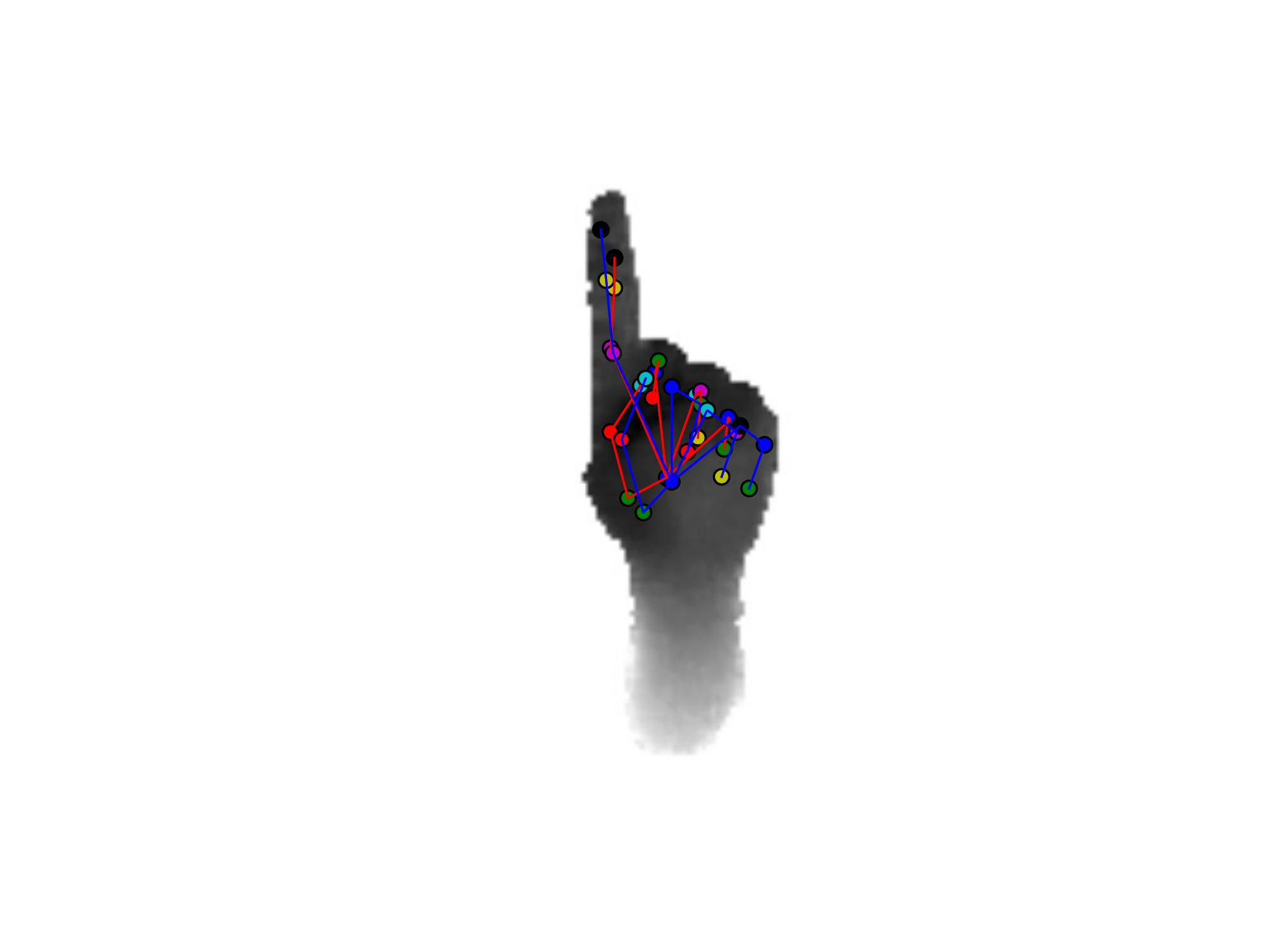} &

\includegraphics[width=0.1\linewidth,trim={5cm 3cm 5cm 3cm},clip]{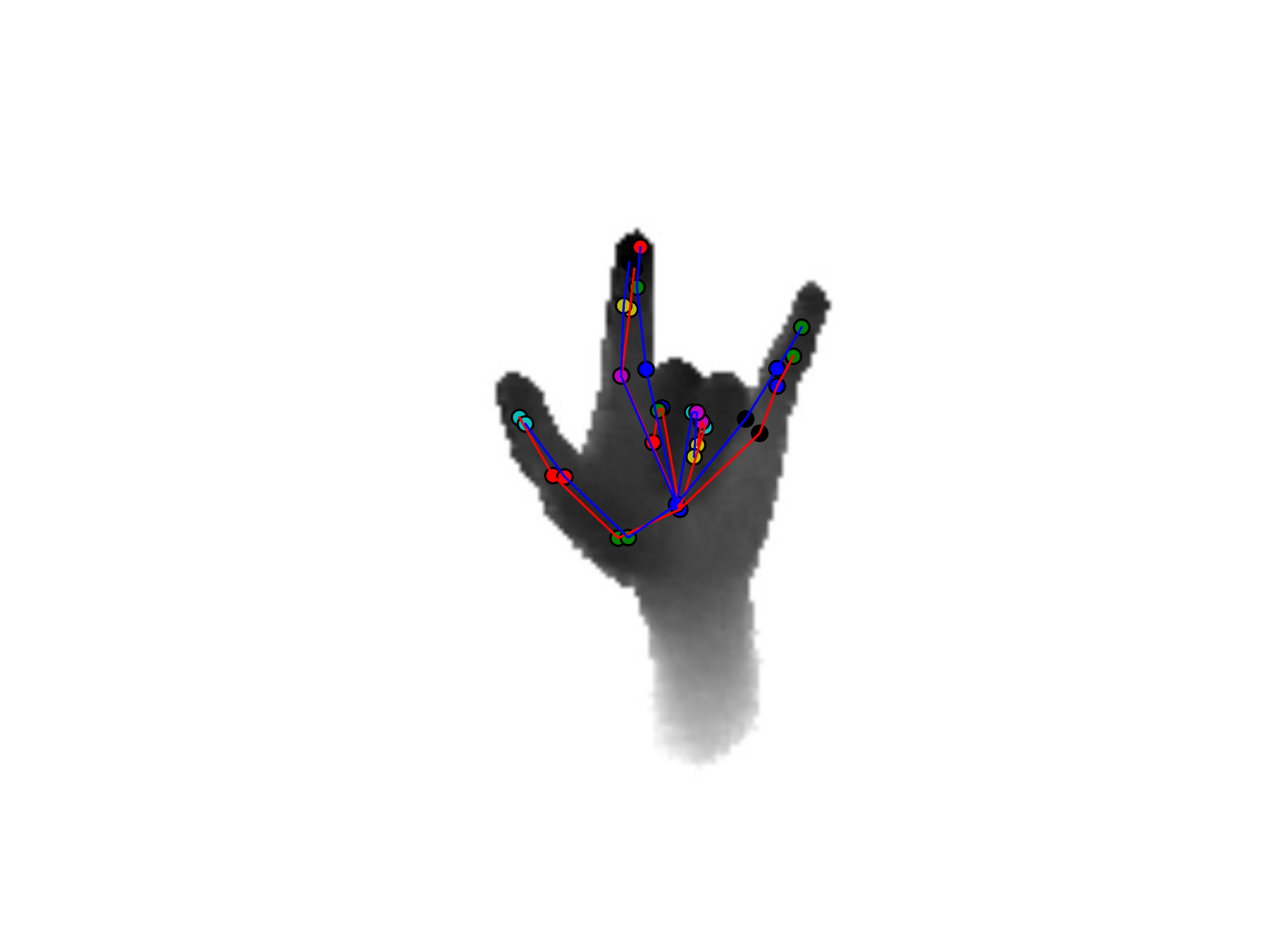} \\

&
\includegraphics[width=0.1\linewidth,trim={2cm 1cm 2cm 1cm},clip]{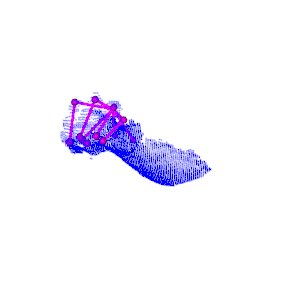} &
\includegraphics[width=0.1\linewidth,trim={1cm 1cm 3cm 1cm},clip]{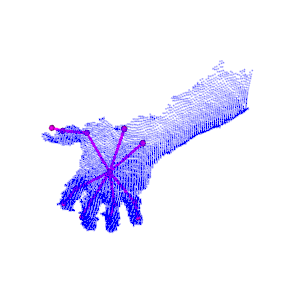} &
\includegraphics[width=0.1\linewidth,trim={1cm 1cm 3cm 1cm},clip]{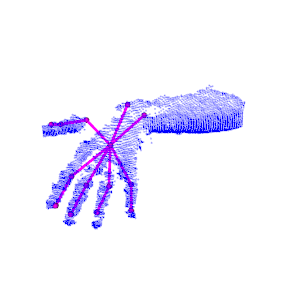} &
\includegraphics[width=0.1\linewidth,trim={2cm 1cm 2cm 1cm},clip]{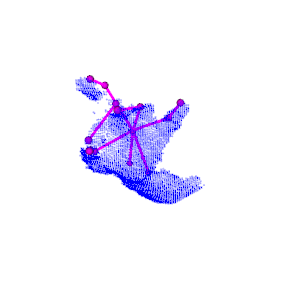} &

\includegraphics[width=0.1\linewidth,trim={2cm 1cm 2cm 1cm},clip]{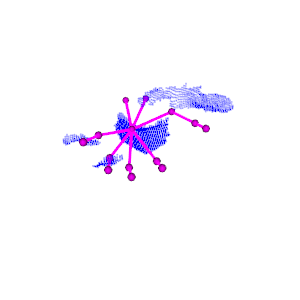} &

\includegraphics[width=0.1\linewidth,trim={2cm 4cm 2cm 1cm},clip]{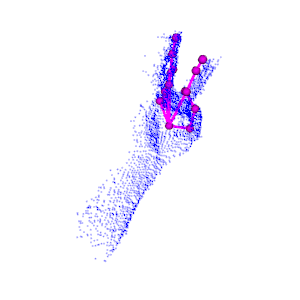} &
\includegraphics[width=0.1\linewidth,trim={2cm 4cm 2cm 1cm},clip]{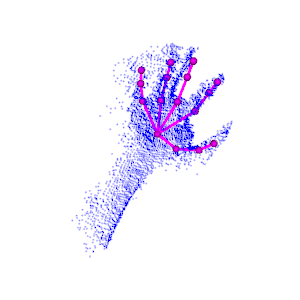} &
\includegraphics[width=0.1\linewidth,trim={2cm 4cm 2cm 1cm},clip]{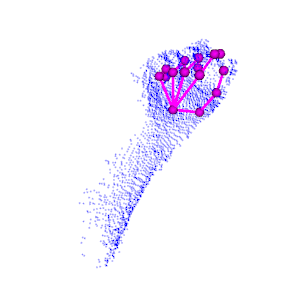} &
\includegraphics[width=0.1\linewidth,trim={2cm 4cm 2cm 1cm},clip]{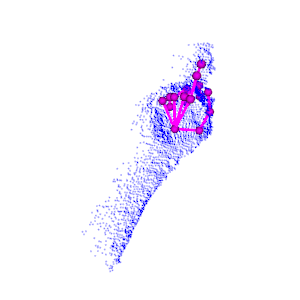} &

\includegraphics[width=0.1\linewidth,trim={2cm 4cm 2cm 1cm},clip]{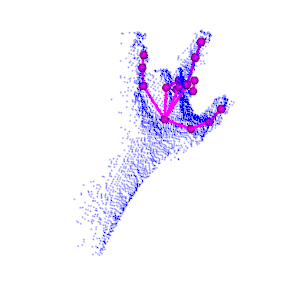} \\

\parbox[t]{3mm}{\multirow{2}{*}{\rotatebox[origin=c]{90}{Deep-ORRef}}} &
\includegraphics[width=0.1\linewidth,trim={5cm 3cm 5cm 3cm},clip]{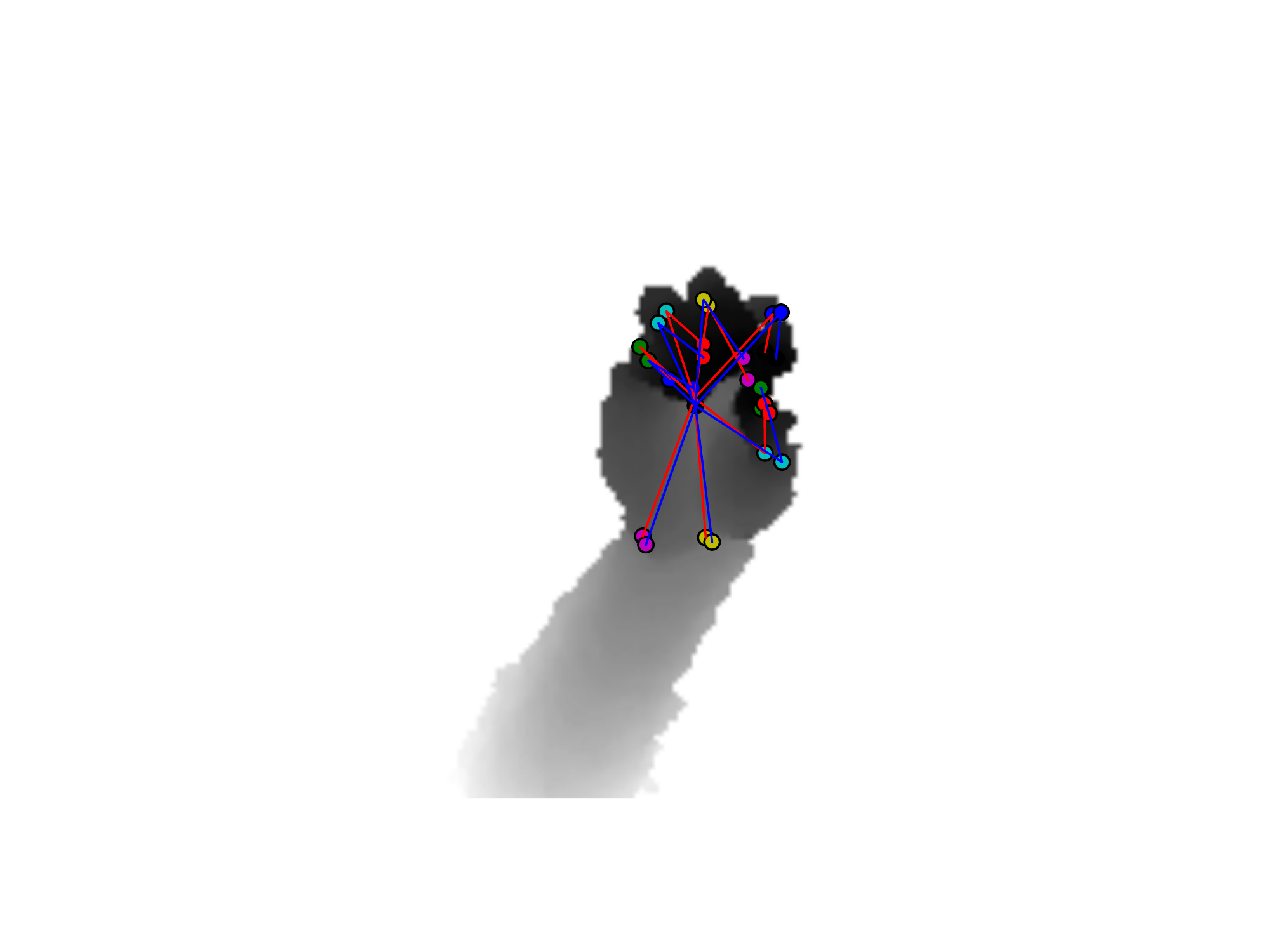} &
\includegraphics[width=0.1\linewidth,trim={5cm 3cm 5cm 3cm},clip]{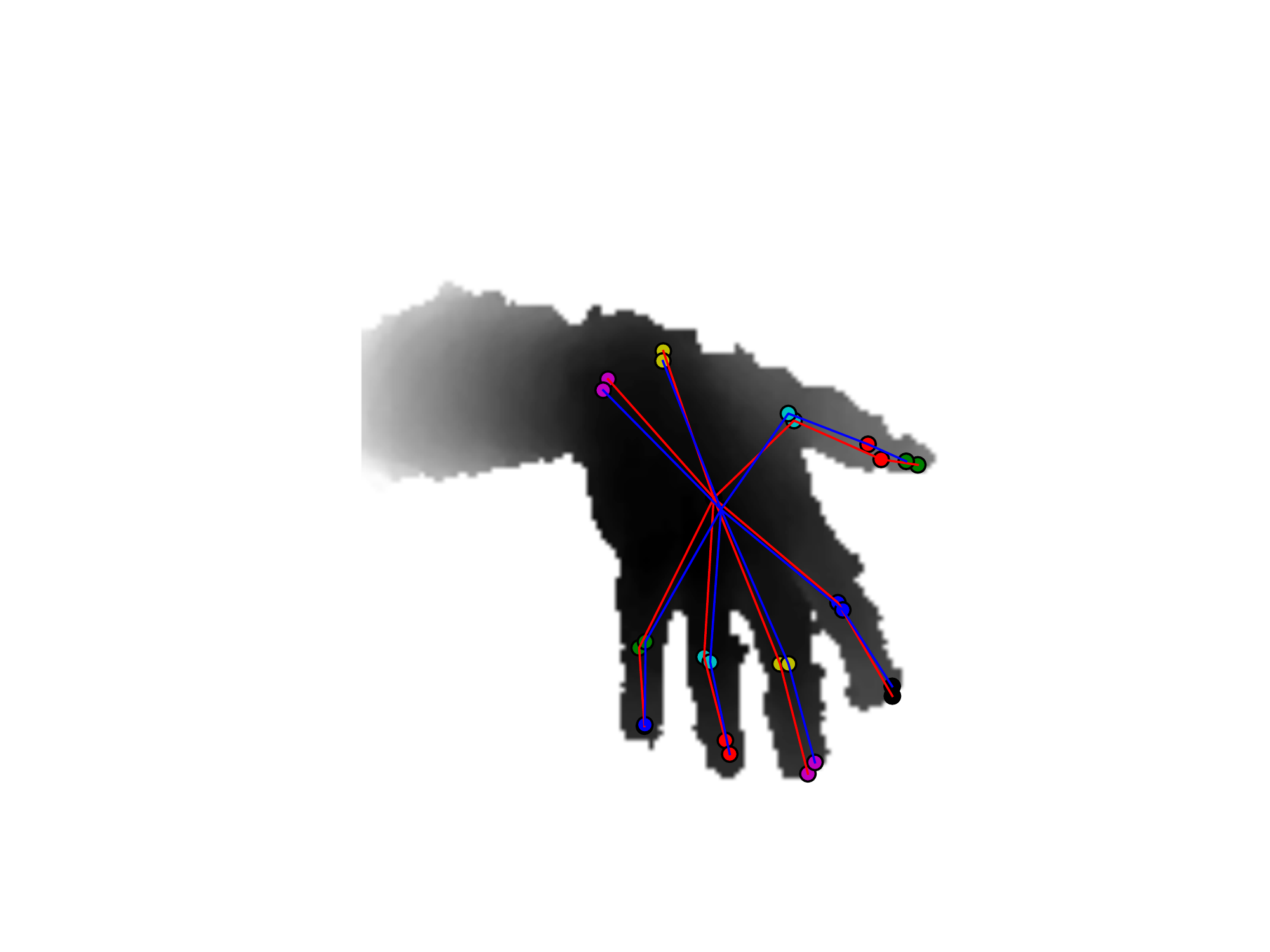} &
\includegraphics[width=0.1\linewidth,trim={5cm 3cm 5cm 3cm},clip]{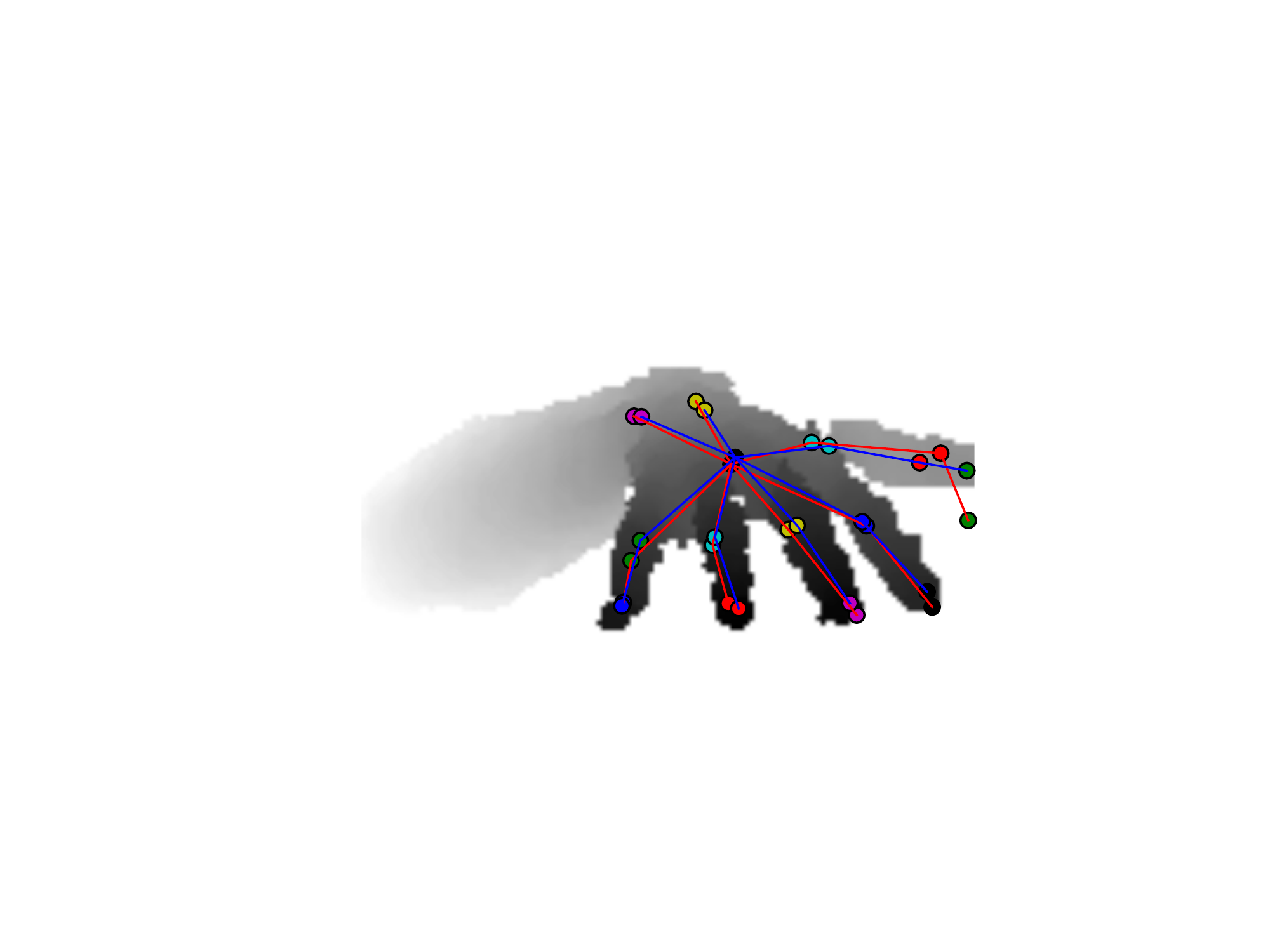} &
\includegraphics[width=0.1\linewidth,trim={5cm 3cm 5cm 3cm},clip]{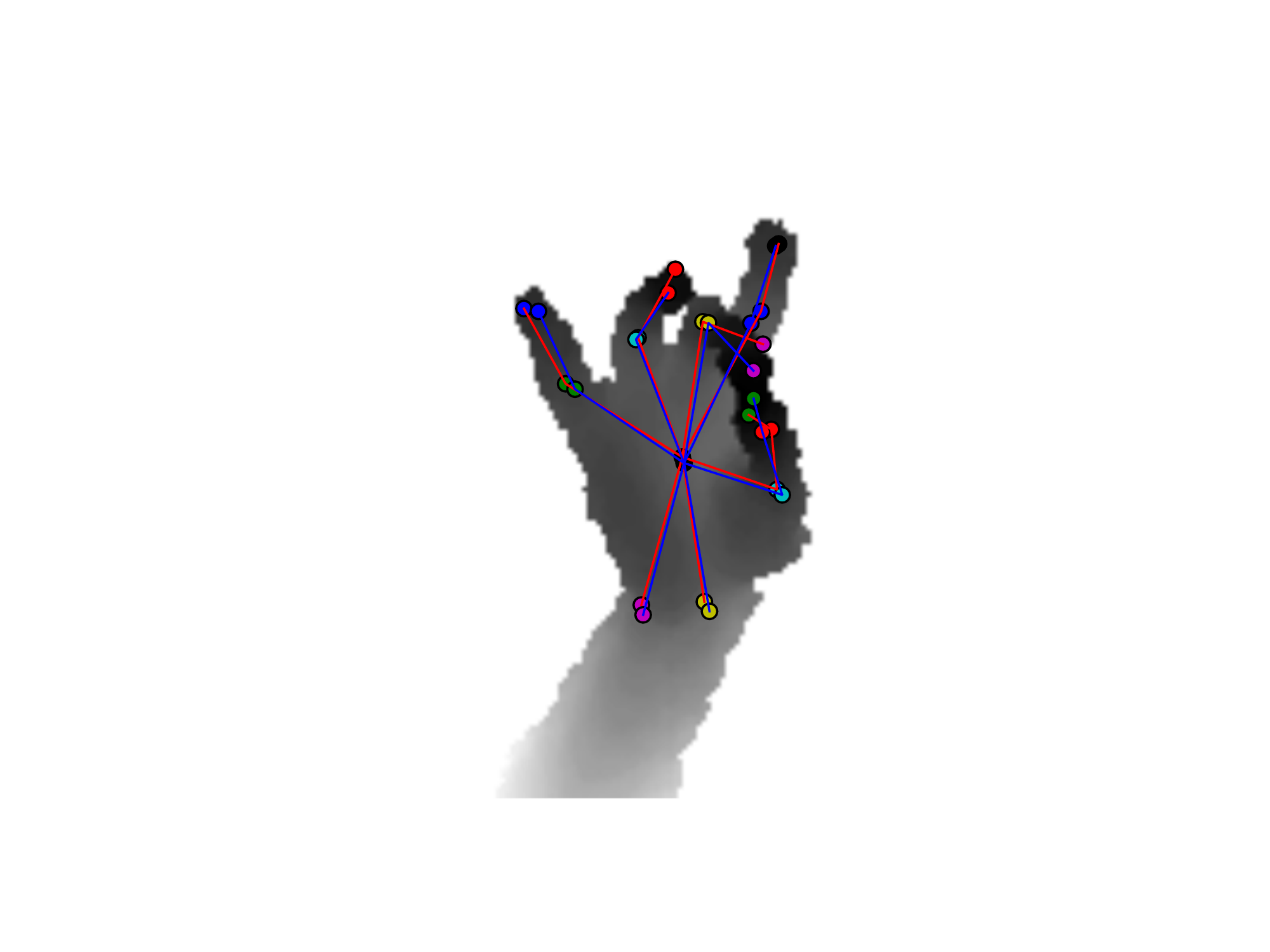} &

\includegraphics[width=0.1\linewidth,trim={5cm 3cm 5cm 3cm},clip]{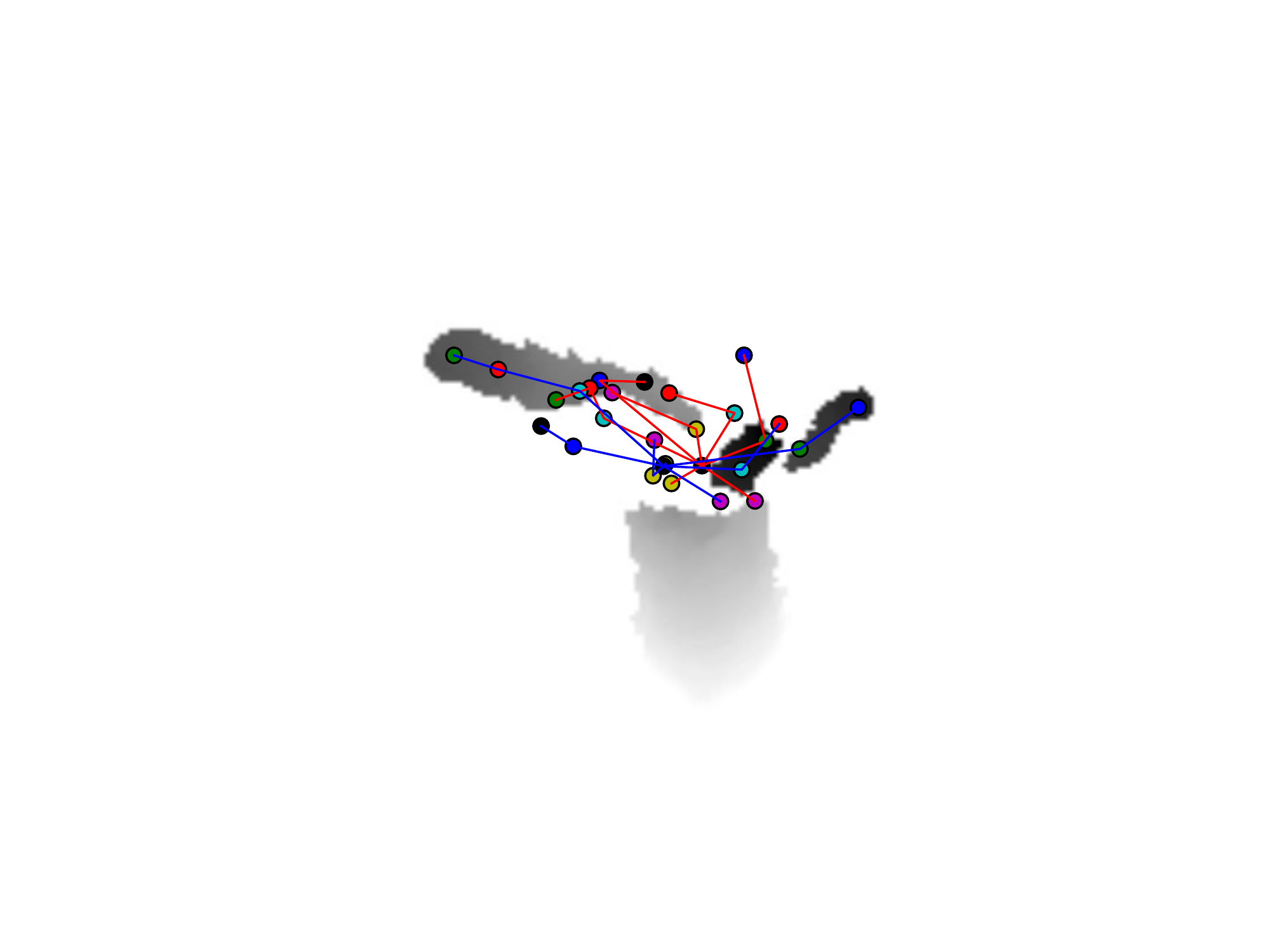} &

\includegraphics[width=0.1\linewidth,trim={5cm 3cm 5cm 3cm},clip]{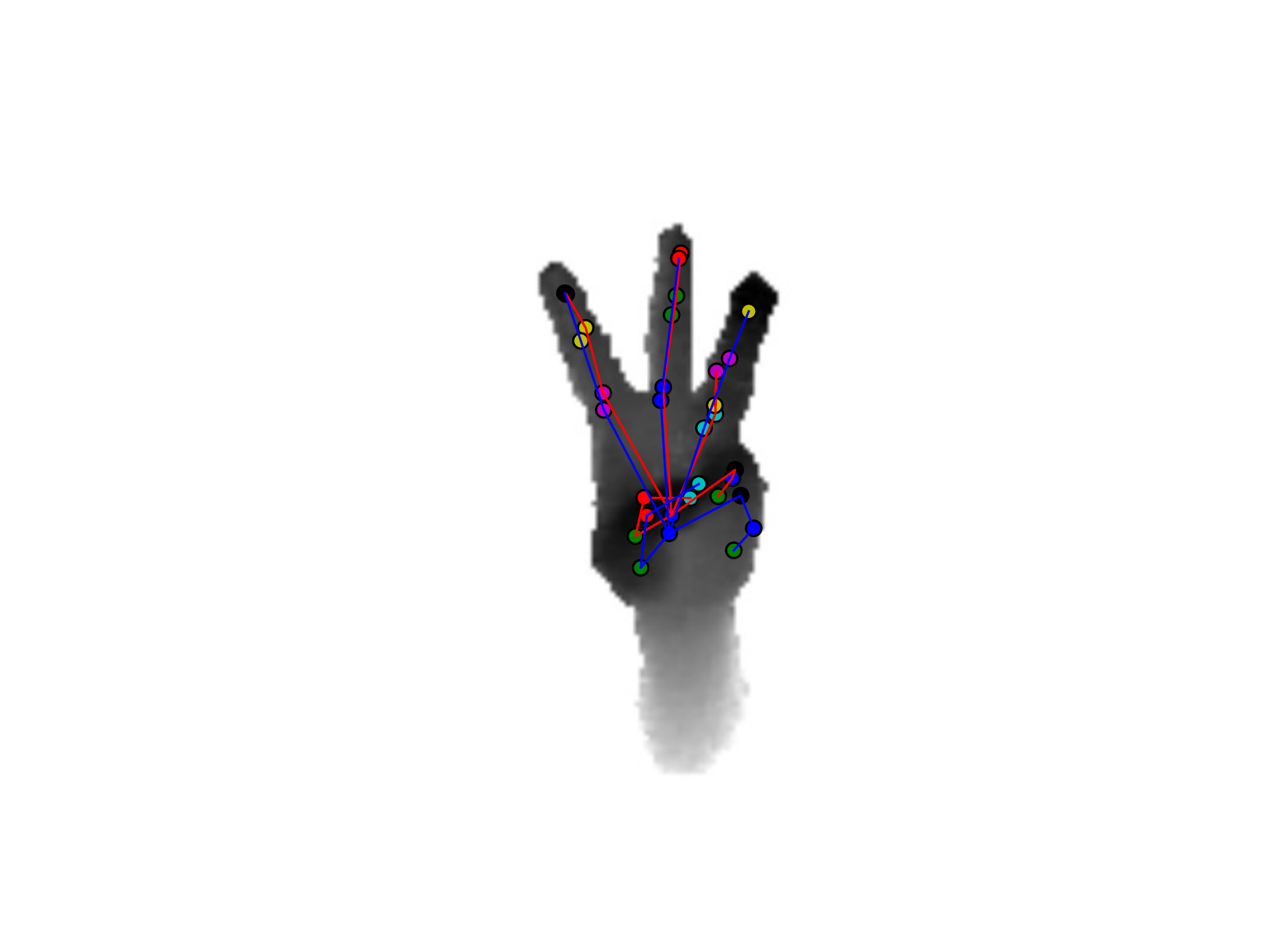} &
\includegraphics[width=0.1\linewidth,trim={5cm 3cm 5cm 3cm},clip]{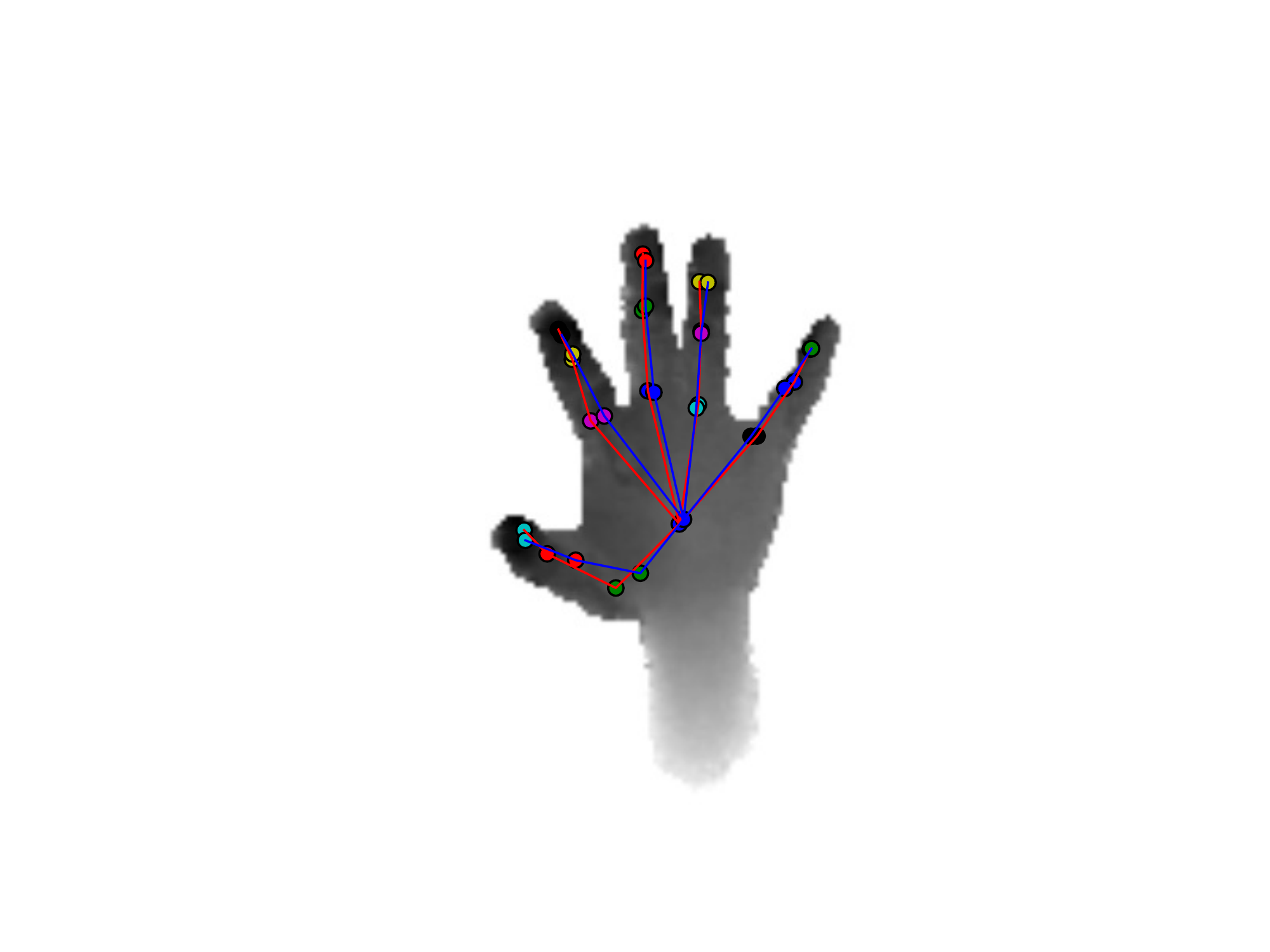} &
\includegraphics[width=0.1\linewidth,trim={5cm 3cm 5cm 3cm},clip]{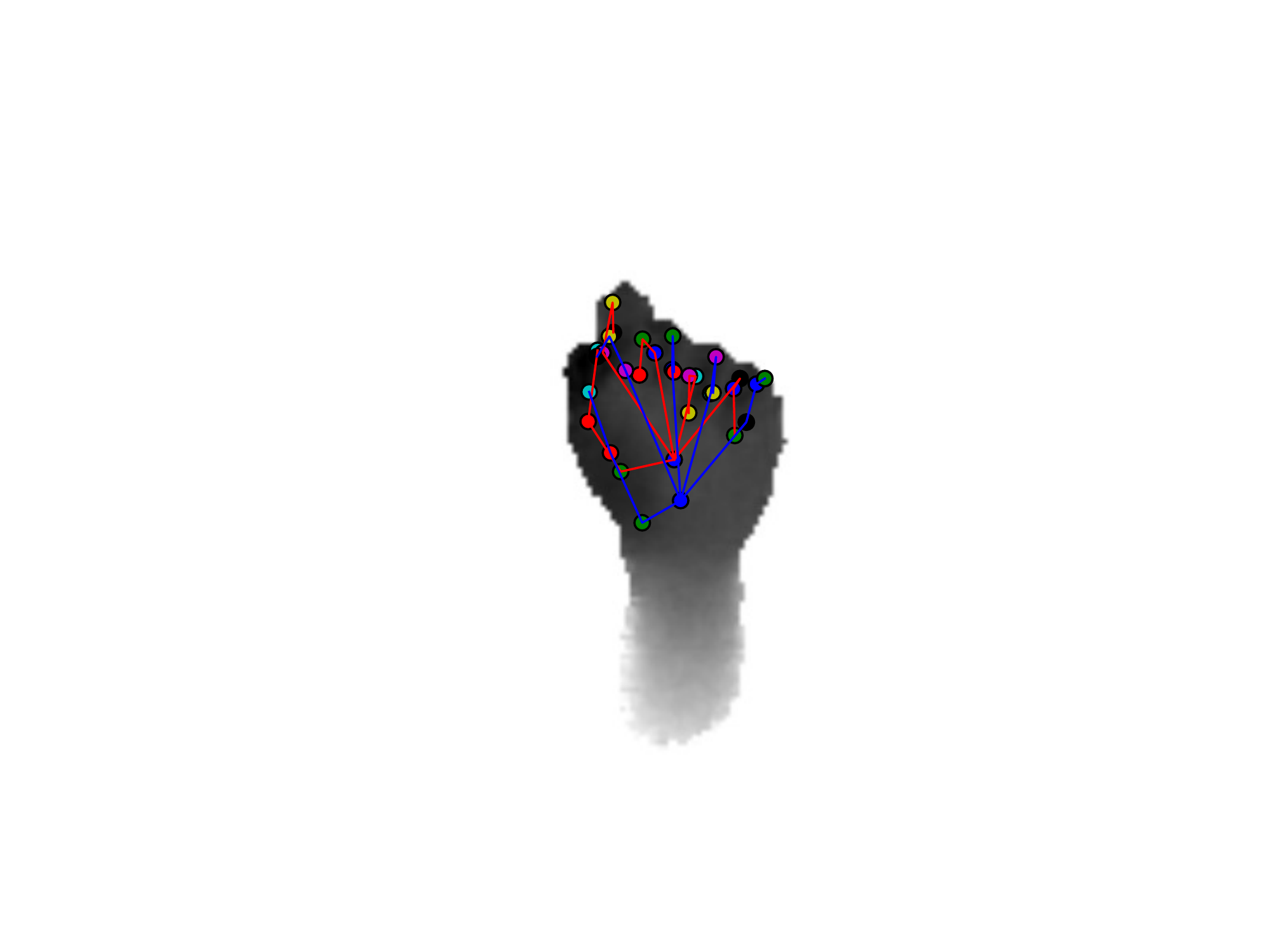} &
\includegraphics[width=0.1\linewidth,trim={5cm 3cm 5cm 3cm},clip]{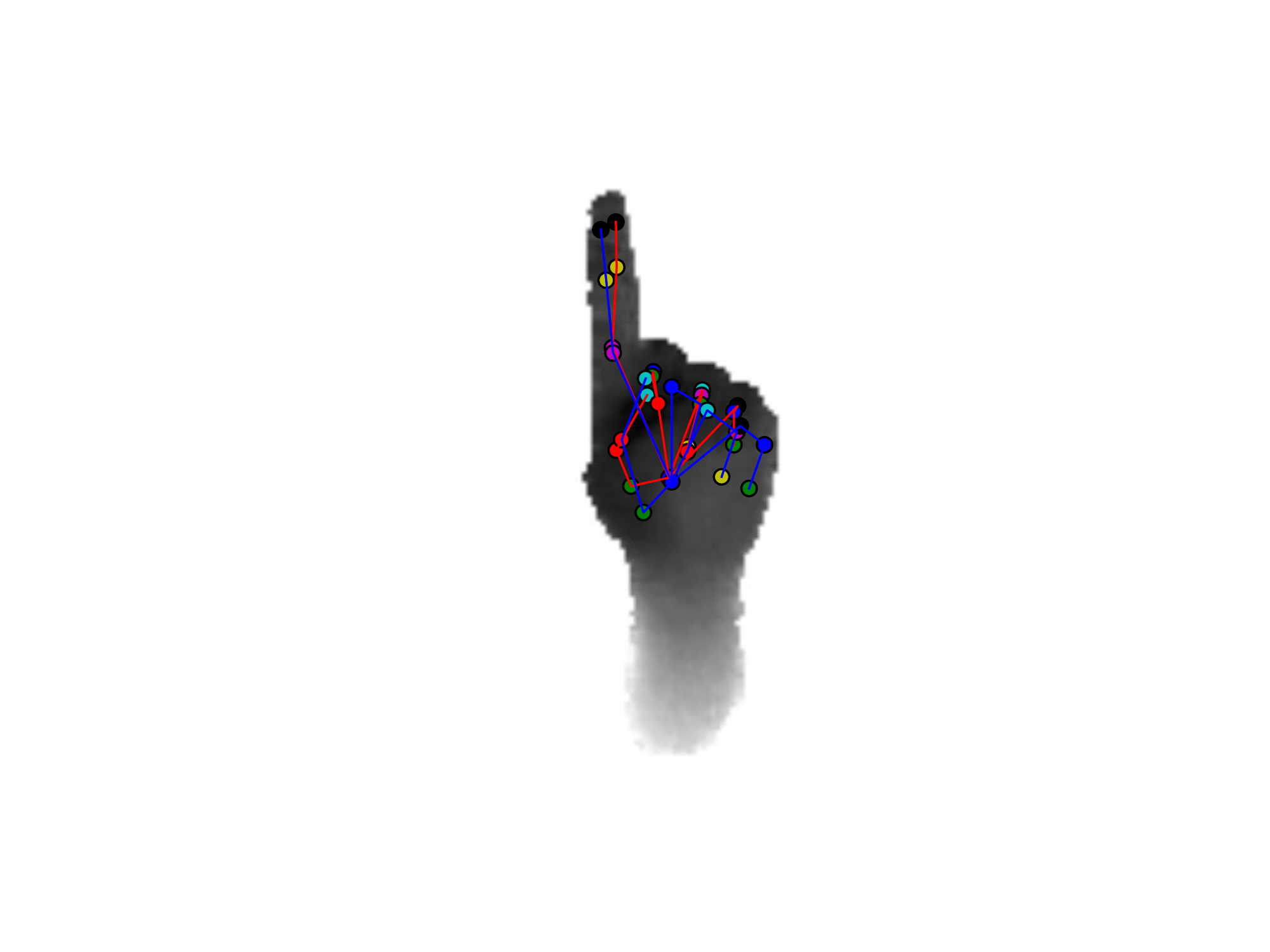} &

\includegraphics[width=0.1\linewidth,trim={5cm 3cm 5cm 3cm},clip]{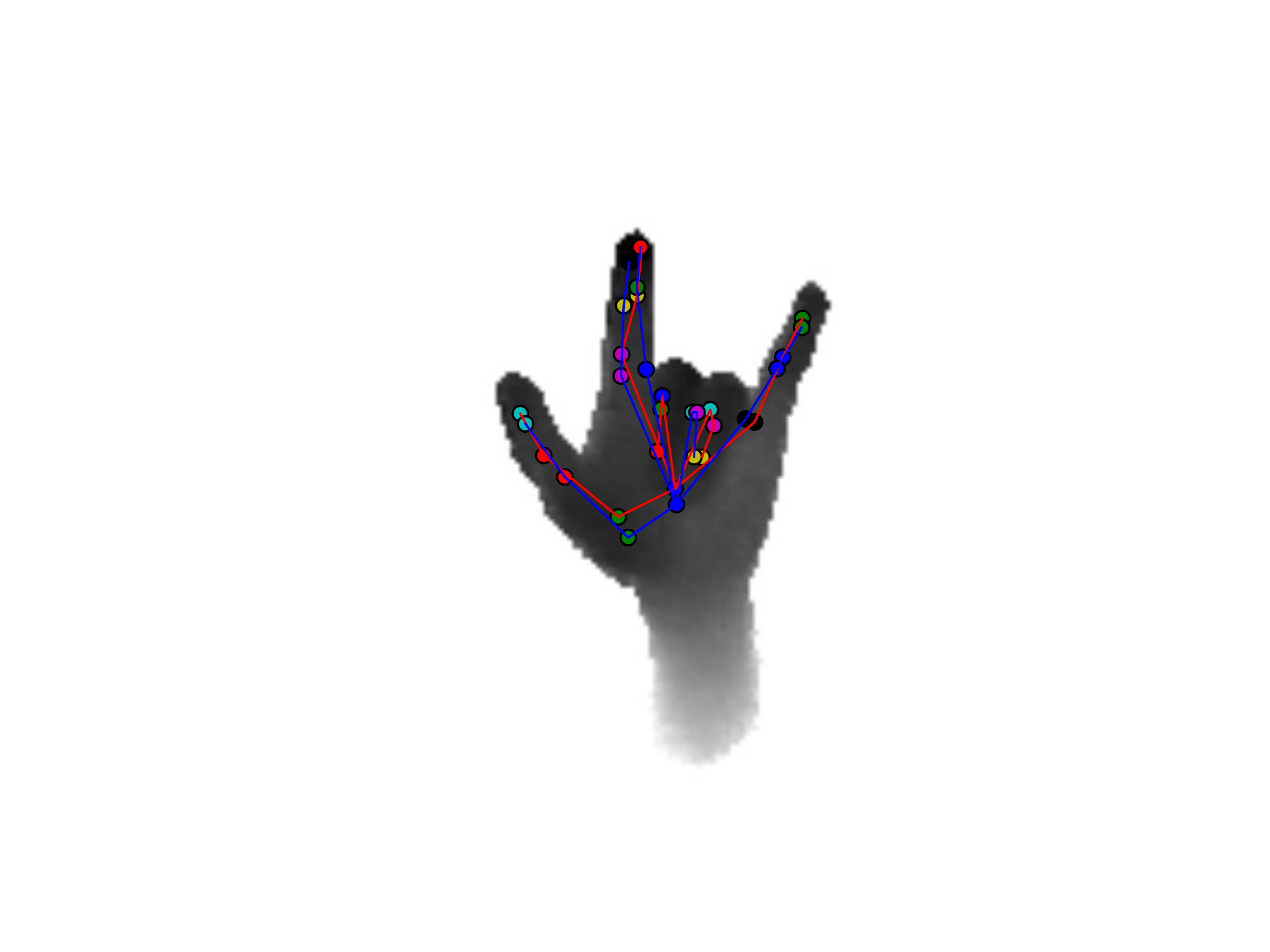} \\

&
\includegraphics[width=0.1\linewidth,trim={2cm 1cm 2cm 1cm},clip]{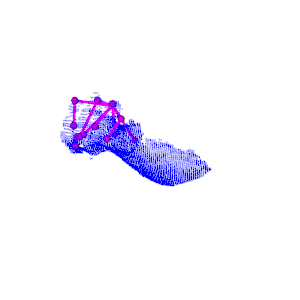} &
\includegraphics[width=0.1\linewidth,trim={1cm 1cm 3cm 1cm},clip]{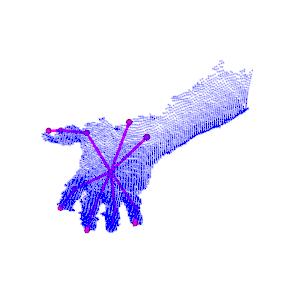} &
\includegraphics[width=0.1\linewidth,trim={1cm 1cm 3cm 1cm},clip]{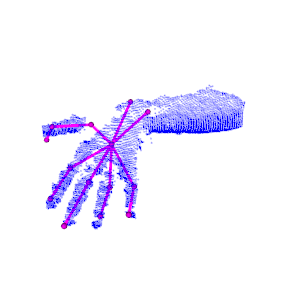} &
\includegraphics[width=0.1\linewidth,trim={2cm 1cm 2cm 1cm},clip]{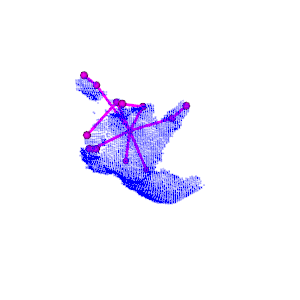} &

\includegraphics[width=0.1\linewidth,trim={2cm 1cm 2cm 1cm},clip]{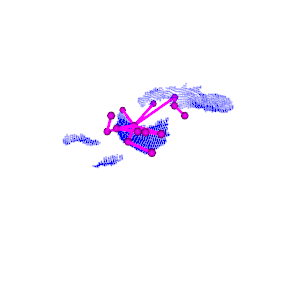} &

\includegraphics[width=0.1\linewidth,trim={2cm 4cm 2cm 1cm},clip]{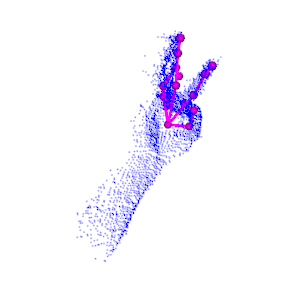} &
\includegraphics[width=0.1\linewidth,trim={2cm 4cm 2cm 1cm},clip]{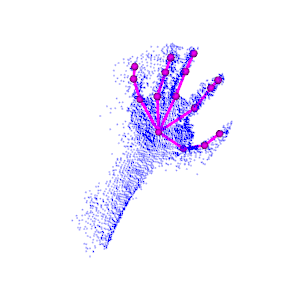} &
\includegraphics[width=0.1\linewidth,trim={2cm 4cm 2cm 1cm},clip]{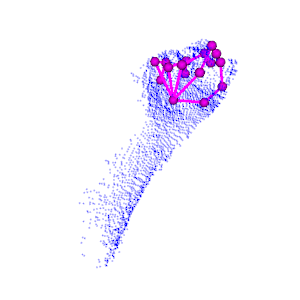} &
\includegraphics[width=0.1\linewidth,trim={2cm 4cm 2cm 1cm},clip]{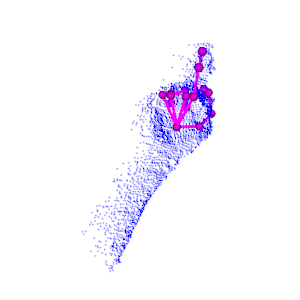} &

\includegraphics[width=0.1\linewidth,trim={2cm 4cm 2cm 1cm},clip]{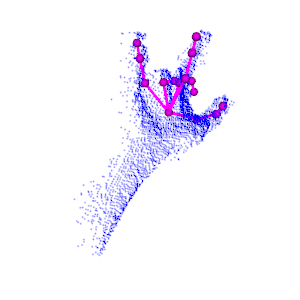} \\

\end{tabular}
\end{center}
   \caption{Qualitative results.   We show the  inferred joint locations  on the
     depth images  (in gray-scale), as well  as the 3D locations  with the point
     cloud of  the hand (blue  images) from a different angle. The ground truth  is shown in  blue, our
     results in  red.  The point  cloud is only  annotated with our  results for
     clarity. The  right columns show  some erroneous  results. One can  see the
     difference between  the global constrained  pose and the  local refinement,
     especially in  the presence of missing  depth values as shown  in the fifth
     column. While the global pose constraint still preserves the hand topology,
     the local refinement cannot reason  about the locations without the missing
     depth data. (Best viewed on screen)}
\label{fig:results_qualitative}
\end{figure*}
\egroup

\section{Conclusion}

We  evaluated  different  network  architectures for  hand  pose  estimation  by
directly regressing the  3D joint locations.  We introduced  a constrained prior
hand   model   that   can   significantly   improve   the   joint   localization
accuracy. Further, we applied a  joint-specific refinement stage to increase the
localization  accuracy. We  have  shown, that  for this  refinement  a CNN  with
overlapping input  patches with different  pooling sizes can benefit  from both,
input resolution and  context.  We have compared  the architectures on two  datasets and
shown that they outperform previous state-of-the-art both in terms of localization
accuracy and speed.

\paragraph*{Acknowledgements:} 
This work was funded by the Christian Doppler Laboratory for Handheld Augmented Reality and the TU Graz FutureLabs fund.

\clearpage

{\small
\bibliographystyle{ieee}
\bibliography{string,b2}
}

\end{document}